\documentclass[11pt]{article}
\usepackage{amsmath,amssymb,amsfonts,amsthm,epsfig}
\usepackage{tikz}
\usetikzlibrary{graphs}
\usepackage{cancel}
\usepackage{fullpage}
\usepackage{liyang}
\usepackage{framed}
\usepackage{verbatim}
\usepackage{enumitem}
\usepackage{array}
\usepackage{multirow}
\usepackage{afterpage}
\usepackage{mathrsfs}

\usepackage[normalem]{ulem}

\newcommand{\ol}[1]{\overline{#1}}

\usepackage{caption} 

\usepackage{todonotes}

\makeatletter
\newtheorem*{rep@theorem}{\rep@title}
\newcommand{\newreptheorem}[2]{
\newenvironment{rep#1}[1]{
 \def\rep@title{#2 \ref{##1}}
 \begin{rep@theorem}\itshape}
 {\end{rep@theorem}}}
\makeatother
\theoremstyle{plain}



\newcommand{\ignore}[1]{}

\def\colorful{1}

\definecolor{darkgreen}{RGB}{10,140,5}

\ifnum\colorful=1

\fi
\ifnum\colorful=0

\fi

\usepackage{boxedminipage}

\newreptheorem{theorem}{Theorem}
\newtheorem*{theorem*}{Theorem}
\newreptheorem{lemma}{Lemma}
\newreptheorem{proposition}{Proposition}
\newtheorem*{noclaim*}{Claim}

\newcommand{\DT}{\mathrm{DT}}

\newcommand{\nco}{\mathsf{NC^1}}

\newcommand{\rank}{\mathrm{rank}}

\newcommand{\CNF}{\mathrm{CNF}} 
\newcommand{\DNF}{\mathrm{DNF}}

\usepackage{dsfont}
\renewcommand{\F}{\mathds{F}}
 
\renewcommand{\R}{\mathds{R}}

\newcommand{\DL}{\mathrm{DL}}

\begin{document}

\title{The Probably Approximately Correct Learning Model in Computational Learning Theory} 
\date{}

\author{Rocco A. Servedio\\
Columbia University}

\maketitle 

\setcounter{page}{1}

\section{Introduction}

Leslie Valiant's 1984 paper ``A Theory of the Learnable'' \cite{Valiant:84}, reproduced in this volume, has the unusual distinction of having changed the course of several scientific disciplines.  Within theoretical computer science it was one of the key works giving rise to the field now known as \emph{computational learning theory}, which may loosely be defined as the rigorous study of learning processes and phenomena from the computer science perspective of efficient algorithms and computational complexity. In the decades since the publication of \cite{Valiant:84}, computational learning theory has grown into a rich field with strong connections to many other theoretical disciplines such as mathematical probability and statistics, information theory, decision theory and more.  Beyond the realm of theory, Valiant's paper and the Probably Approximately Correct (PAC) model which he introduced in it have also had a great impact on the subsequent development of \emph{machine learning}, a field which has already transformed many aspects of science and human society and seems certain to have an even greater influence in the future. 

This chapter gives an overview of the Probably Approximately Correct learning model that Valiant introduced in \cite{Valiant:84}, explaining some of the major results and directions that the field has taken in the years since that work.   As we shall see, a foundational research program naturally emerges from the (remarkably elegant) basic framework of PAC learning given in \cite{Valiant:84}, which is to delineate the boundary between what is and what is not learnable by computationally efficient agents. Indeed, as described below, this is one of the chief scientific goals which is expressly advocated in Valiant's original paper.  A particular focus of this chapter is to summarize and explain some of the (considerable) progress that has been made towards this goal. 

  As we shall see, the research community's understanding of efficient PAC learnability has come a long way since the publication of \cite{Valiant:84}.  However, a recurring theme which emerges throughout this survey is that Valiant's work on PAC learning in the 1980s --- including \cite{Valiant:84} and \cite{KearnsValiant:94}, both of which are included in this volume, and other works --- was remarkably prescient in anticipating major themes and directions which subsequently grew into substantial strands of research for the computational learning theory community.  Another theme is the interplay of ideas and techniques across multiple domains within (and beyond) theoretical computer science, spanning domains such as statistics, information theory, algorithms, complexity, cryptography, and more, which recur throughout the study of efficient learnability and lend the field of computational learning theory much of its intellectual richness.

\medskip

{\bf Organization.}
The structure of this article is as follows.  In~\Cref{sec:framework} we define the basic ``distribution-free'' PAC learning framework which was introduced by Valiant in \cite{Valiant:84} and highlight the central question which this framework gives rise to (and which is the main topic of this article):  

\begin{quote}
Which functions are learnable by computationally efficient algorithms, and which are not?
\end{quote}

\Cref{sec:algorithms} gives an overview of positive results, namely efficient PAC learning algorithms for various important classes of functions in Valiant's original distribution-free framework as well as in a number of variants of the basic framework. The models we consider include ones with and without black-box queries, with and without noise corrupting the labels of examples, and distribution-specific variants of the original distribution-independent framework.  As we shall see, a wide range of algorithmic ideas, involving combinatorial and probabilistic reasoning as well as tools from analysis, geometry, algebra, probability, and other areas, have been employed in developing efficient learning algorithms in these models. 

Complementing these positive results,~\Cref{sec:hardness} surveys \emph{computational hardness} results for PAC learning.  These are rigorous theoretical statements establishing formal limitations on what can be PAC learned by efficient algorithms, based on the presumed intractability of various computational problems.
That section briefly describes some \emph{representation-dependent} hardness results, which establish computational hardness for PAC learning algorithms that use a particular form of hypothesis representation and rely on worst-case hardness assumptions of the P $\neq$ NP flavor. However, most of the emphasis is on \emph{representation-independent} results, which show that certain types of functions are \emph{inherently unpredictable}; these kinds of results rely on stronger average-case (sometimes called ``cryptographic'') hardness assumptions. Hardness results of the latter sort are of particular interest, as they very directly capture a strong sense in which computational hardness limits learning. These types of results were first developed in the paper ``Cryptographic Limitations on Learning Boolean Formulae and Finite Automata'' by Kearns and Valiant \cite{KearnsValiant:94}, which is included in this volume.

\section{The PAC Learning Framework} \label{sec:framework}

We begin this section by giving some intuition and background for the general PAC learning framework in~\Cref{sec:motivation}\ignore{our exposition is relatively concise since the aim is to pave the way to more technical material}.  Building on this intuition,~\Cref{sec:definitions} gives a more detailed description of the key ingredients of the PAC learning model.  As an illustrative example, a simple PAC learning algorithm is presented there for learning a basic concept class (the class of all Boolean conjunctions), along with a sketch of its analysis.  With this foundation in place, \Cref{sec:program} explains the research program of determining which classes of Boolean functions are efficiently learnable, which is the main subject of this survey.

\subsection{Motivation and intuition\ignore{for the PAC learning framework}.}  
\label{sec:motivation}

This subsection concisely  introduces the basic ideas of the PAC learning framework.  A more leisurely treatment, which situates PAC learning theory in a broader context of connections to theories of computation and evolution, can be found in \cite{ValiantPAC}, in particular Chapter~5.

To motivate the general setup of the PAC learning model, it is useful to consider a simple object recognition scenario.  Imagine that a learning agent (such as a small child) finds itself in a world with various kinds of objects (such as illustrations of animals in a picture book), with the goal of learning to recognize a certain type of object given access to objects which have been labeled identifying each one as either a positive or negative example of the target object type. (For example, for each picture that is encountered an adult may tell the child whether or not it is a picture of a dog.)  This is a natural scenario, occurring in some form countless times every day in the real world, and it is clear from common experience that effective and efficient learning is indeed possible.
The PAC learning model offers an abstraction that allows quantitative analysis of the abilities and limitations of computationally bounded learning agents in situations of this sort.

Let us recast the above scenario using some mathematical terminology.  The \emph{domain} or \emph{universe} of all possible objects (commonly called \emph{instances}) is a set which we denote by $X$, and the \emph{classification rule} to be learned (often referred to as the \emph{target function} or \emph{target concept}) is a function $f: X \to \zo$ (we may equivalently consider the subset $f^{-1}(1) \subseteq X$ of all instances which are labeled positively by $f$).  We assume that there is an unknown and arbitrary probability distribution, denoted ${\cal D}$, over the set $X$ of unlabeled instances. A learning algorithm is given access to independent random (instance, label) pairs $(\bx,f(\bx)) \in X \times \zo$, where in each pair the instance $\bx$ is distributed according to ${\cal D}$.\footnote{Throughout this survey we use bold font such as $\bx$ to indicate random variables.}  The goal of a learning algorithm is to output a hypothesis function $h: X \to \zo$ --- intuitively, representing the learner's ``best guess'' at the classification rule $f$ --- which is Probably (i.e.~with probability $1-\delta$ for some small \emph{confidence parameter} $\delta$) Approximately Correct. Here ``Approximately Correct'' means that for some small \emph{error parameter} $\eps > 0$, the hypothesis $h$ satisfies
\[
\Prx_{\bx \sim {\cal D}}[h(\bx) \neq f(\bx)] \leq \eps.
\]
Note that a learning algorithm which can output an approximately correct hypothesis $h$ can use that hypothesis $h$ to make high-accuracy predictions on ``future'' examples drawn from ${\cal D}$; indeed, the ability to make high-accuracy predictions implicitly defines a hypothesis function that the learning agent is using to make its  predictions.

To motivate the above $(\eps,\delta)$-criterion for successful learnability, we note that in general it may be impossible for a learning algorithm to efficiently output an exactly correct hypothesis, since doing so may require exhaustive exposure to all possible instances to rule out very rare ``corner cases''.  Moreover, in a probabilistic model such as the one described above, even outputting an approximately correct hypothesis cannot be guaranteed with absolute certainty since it is conceivable that an unfortunate learning algorithm may only be exposed to a very unrepresentative sample of labeled examples because of suffering from poor ``luck of the draw''. Thus both the $\eps$ and $\delta$ parameters are necessary given the probabilistic scenario that we consider; fortunately, we shall see that both these sources of error $\eps$ and $\delta$ can be made arbitrarily small.

The PAC learning framework has as a central concern both the \emph{running time} and \emph{sample complexity} (number of labeled examples) required for a learning algorithm to output its hypothesis (as well as the running time necessary to evaluate the hypothesis function $h$; we discuss this further in~\Cref{sec:definitions}).  Finally, a standard learning problem in the PAC setting does not deal with simply learning a single unknown concept $f$; rather, there is an a priori known-to-the-learner \emph{concept class} ${\cal C}$ of possible target functions (typically characterized by some type of simple syntactic Boolean circuit structure that is shared by all the functions in ${\cal C}$), with the promise that the actual unknown target function $f$ belongs to ${\cal C}$.

We briefly discuss some of the intuition behind the basic model sketched above.  The unrestricted distribution ${\cal D}$ over the instance space $X$ captures the notion that in general little may be known about the settings in which nature could require a learning algorithm to perform; by requiring our learning algorithms to work for arbitrary distributions ${\cal D}$, the PAC model minimizes assumptions about the environment.  Having examples for the learner be independently generated according to the fixed distribution ${\cal D}$, and assessing the accuracy of the learned hypothesis $h$ with respect to the same distribution ${\cal D}$, corresponds to the now-standard ``train-and-test'' paradigm in machine learning.  More fundamentally, this aligns with the natural intuition that learned experience may only be useful in an environment similar or identical to the one in which the learning took place (if an animal has learned which plants are good to eat in the savannah but finds itself in a rain forest, its learning may not be of use).  Finally, there are several reasons for the assumption that instances are labeled according to a target function $f$ which belongs to some class ${\cal C}$ of syntactically simple functions.  One is that some regularity or ``signal in the data'' must be present in order for any meaningful type of learning to be possible --- if each successive example were independently labeled by a fresh fair coin toss rather than by a fixed Boolean function $f$, then clearly it would be impossible to learn any useful predictor for the labels of future examples.  Another is the fact that we are interested in the learning abilities of  \emph{computationally} (running time) and \emph{statistically} (amount of data required) bounded agents; in particular, computational limitations of the learner turn out to necessitate the assumption that the underlying classification rule or target function is ``not too complex.'' 
As is discussed further in \Cref{sec:program}, the fundamental goal of research in this area is to understand ``how complex'' are the functions that can be efficiently learned; this complexity of functions is naturally captured by a concept class ${\cal C}$.  
Indeed, understanding the complexity of Boolean functions by studying associated classes of Boolean circuits is a longstanding and standard approach taken in the field of computational complexity theory.
As we shall see, a central broad message which emerges from the line of work on PAC learning which is surveyed in this chapter is that while some simple functions can be learned efficiently,  more complex functions often cannot.  

From a modern point of view the conceptual framework sketched above may seem natural or even obvious, but we remark that this is an advantage bestowed by several decades of hindsight.  Prior to Valiant's definition of the PAC learning model,  there had been various efforts to formalize the study of learning from different perspectives:  for example, an ``inductive inference'' model of learning grounded in computability theory had been proposed by Gold \cite{Gold:67}, and a statistical model of learning was given by Vapnik \cite{Vapnik:82,Vapnik:98}.  But in general, in the early 1980s having a rigorous framework for the study of machine learning problems was very far from being a mainstream idea; in particular, a highly novel and impactful aspect of the PAC model was the idea that both computational \emph{and} statistical considerations must play central roles in an understanding of effective learnability.  The further insight that both aspects can be studied in a quantitatively precise way led to a rich theory; Valiant's framework hit a sweet spot between modeling important aspects of real learning scenarios and capturing fundamental theoretical issues of statistical and algorithmic efficiency in learning.

Before entering into a more technical treatment of PAC learning, we note that the version of the PAC learning model which we have informally described above and which we present and analyze in more detail below does not align perfectly with the discussion in \cite{Valiant:84}.  The original paper  \cite{Valiant:84} additionally considers variants of the model in which learning algorithms may receive only \emph{partial} examples (if each instance in $X$ is an $n$-bit string $x \in \zo^n$, the learning algorithm may receive a string in $\{0,1,\ast\}^n$ where each $\ast$ coordinate serves as a ``don't know'' symbol); it also analyzes a range of oracles (sources of information about the target function) beyond just the random examples described above.  Some subsequent work has considered settings of learning from partial information in PAC-type frameworks (see for example  \cite{BenDavidDichterman:98,BDJ+:98,OS11:chow,DDFS14}), but a richer and more fully developed theory has emerged for the full information model and consequently that model  will be the focus of our discussion.  Later in this article we will consider different kinds of oracles beyond just random example oracles, including some of the ones that Valiant proposed in \cite{Valiant:84}.

\subsection{The distribution-free PAC learning model and some variants.} 
\label{sec:definitions}

Motivated by the intuitive discussion provided in the previous subsection, we now give a more detailed description of the  PAC learning model.  In this subsection we first focus on defining the so-called ``distribution-independent'' model of PAC learning from random examples, and then sketch how this basic model can be altered and extended in various ways (including some extensions already proposed for study by Valiant in \cite{Valiant:84}). An excellent source giving detailed definitions of the PAC model is \cite{KearnsVazirani:94}.

\subsubsection{Key ingredients of the PAC framework.} \label{sec:key-ingredients}

{\bf The instance space.}  As stated earlier there is a \emph{universe} or \emph{instance space} $X$ of possible \emph{instances}; this is the domain for all of the functions which we will think about learning.  In principle $X$ can be any set, but we will generally consider either the domain of all $n$-bit strings, i.e.~$X = \zo^n$ (or equivalently $X=\bn$), or the domain of all $n$-dimensional real vectors, i.e.~$X=\R^n$.  Here ``$n$'' is an asymptotic parameter, and we will be centrally concerned with how the running time and sample complexity of our learning algorithms scale as functions of $n$. 

Following standard practice, when we consider $X=\zo^n$ we typically view the value 1 as corresponding to ``true'' and 0 as corresponding to ``false'' (when we consider $X=\bn$  we view $-1$ as true and $1$ as false).  It is helpful to think of the set $X$ as consisting of all possible representations of objects, where  we think of a given object as being described by a vector of $n$ Boolean  or numerical attributes.  
For example, the first bit of the vector describing a given object might encode whether or not the object is red; the second bit of the vector might encode whether or not the object is made of metal; and so on.

\medskip

\noindent {\bf Concepts and concept classes.}  A \emph{concept} $f$ over instance space $X$ is simply a Boolean-valued function over the domain $X$, i.e a function mapping $X$ to $\zo$ (or to $\bits$; this will sometimes be more convenient).  A concept can equivalently be viewed as a subset of $X$\ignore{, namely the set $f^{-1}(1).$}.  

A \emph{concept class} ${\cal C}$ over instance space $X$ is simply a set of concepts over $X$.  In principle a concept class can be any collection of concepts, but we will most commonly be interested in concept classes which correspond to the set of all concepts that can be expressed using some particular type of 
syntactic representation.  A paradigmatic example is the concept class ${\cal C}$ over $X=\zo^n$ of all functions which are representable as \emph{Boolean conjunctions}. In this example ${\cal C}$ consists of the $3^n$ functions from $\zo^n$ to $\zo$ which are of the form $\ell_{i_1} \wedge \cdots \wedge \ell_{i_j}$, where each $\ell_i$ is either the Boolean literal $x_i$ or the literal $\overline{x}_i$ and $1 \leq i_1 < \cdots < i_j \leq n$. (We will give a simple and efficient PAC learning algorithm for this concept class in \Cref{sec:PAC-learning-example}.)   We will often refer to a concept class simply by referring to the representations of the concepts in the class; for example, we will refer to ``the concept class of all conjunctions'' as shorthand for ``the concept class of all functions which are representable as Boolean conjunctions.''

\medskip

\noindent {\bf Distributions, target concepts, and example oracles.}
We write ${\cal D}$ to denote a probability distribution over the instance space $X$.  Unless otherwise specified, the distribution ${\cal D}$ may be completely arbitrary.

In the PAC model of learning a particular concept class ${\cal C}$, the learning algorithm is trying to learn an unknown \emph{target concept}, which is a function $f$ that is guaranteed to belong to ${\cal C}$ but is otherwise unknown to the learner.  The learner has access to an \emph{example oracle}, which is a source of labeled examples. The example oracle, which we denote by $EX(f,{\cal D})$, takes no input, and each time it is invoked it outputs a pair $(\bx,f(\bx)) \in X \times \zo$, where $\bx$ is independently drawn from ${\cal D}$ and $f$ is the target concept. (We stress that the learning algorithm does not know the identity of $f$ within ${\cal C}$, and that the distribution ${\cal D}$ is also unknown to the learner.).  

\medskip

\noindent {\bf Hypotheses and hypothesis classes.}
In the PAC model a learning algorithm makes some number $m$ of draws from $EX(f,{\cal D})$ to obtain a \emph{labeled sample} $(x^1,f(x^1)), \dots, (x^m,f(x^m))$; then it does some computation and outputs a \emph{hypothesis} $h$.
 A \emph{hypothesis} is a representation of a Boolean function from $X$ to $\zo$ (or to $\bits$); if a learning algorithm for a concept class ${\cal C}$ always outputs a hypothesis from a particular class ${\cal H}$ of hypothesis representations, we say that the algorithm \emph{learns ${\cal C}$ using ${\cal H}$.}  
 We will generally be concerned with hypothesis classes consisting of those hypotheses that have some specific syntactic structure (such as the hypothesis class of all Boolean conjunctions). Later we will also consider the computational complexity of {evaluating} different types of hypothesis representations.
 
 The \emph{error} of a hypothesis $h$ on target concept $f$ under distribution ${\cal D}$ is defined as
 \[
\text{error}_{\cal D}(h,f) :=  \Prx_{\bx \sim {\cal D}}[h(\bx) \neq f(\bx)].
\]

Now we can give a precise definition of Probably Approximately Correct learning, formalizing the intuition from \Cref{sec:motivation}.  Fix the instance space to be either $X=\zo^n$ or $X=\R^n$, and fix a concept class ${\cal C}$ over $X$.  An algorithm $A$ is said to be a \emph{PAC learning algorithm for ${\cal C}$ using hypothesis class ${\cal H}$} if there are functions $m(n,\eps,\delta)$ and $t(n,\eps,\delta)$ such that the following holds:  Let ${\cal D}$ be any distribution over $X$ and let $f$ be any target concept in ${\cal C}$. Let $0<\eps,\delta<1$ be any \emph{accuracy} and \emph{confidence} parameters.  Given access to the example oracle $EX(f,{\cal D})$ and given the parameters $\eps$ and $\delta$, the algorithm $A$ makes at most $m(n,\eps,\delta)$ calls to $EX(f,{\cal D})$, runs for time at most $t(n,\eps,\delta),$ and with probability at least $1-\delta$ it outputs a hypothesis representation $h \in {\cal H}$ satisfying $\text{error}_{\cal D}(h,f) \leq \eps.$
(The time bound $t(n,\eps,\delta)$ is often informally referred to  as the \emph{running time} of the learning algorithm, and the number of oracles calls $m(n,\eps,\delta)$ is referred to as its \emph{sample complexity}.)

\medskip

\noindent
{\bf Discussion, extensions, and variants.}    We will often be concerned with concept classes whose definition involves a \emph{size parameter}, such as the class ${\cal C}$ over $\zo^n$ of all functions expressible as size-$s$ DNF formulas.  (A DNF formula is an OR of Boolean conjunctions, which are often called \emph{terms} in this context, and the \emph{size} of a DNF is the number of terms that are ORed together in it.) In such settings the running time and sample complexity will also depend on the size parameter $s$.  

As stated earlier, the paramount goal in the PAC learning model is to obtain PAC learning algorithms, for various concept classes, which are efficient both in terms of sample complexity and running time.  Sometimes it is of particular interest to have PAC learning algorithms in which the hypothesis class ${\cal H}$ coincides with the concept class ${\cal C}$ (or more precisely, in which every hypothesis representation $h \in {\cal H}$ represents a function in ${\cal C}$); algorithms of this sort are referred to as \emph{proper} PAC learning algorithms.  We will also be interested in non-proper PAC learning algorithms, for which ${\cal H}$ need not align with ${\cal C}$; however, one important restriction on ${\cal H}$ is that it would not be reasonable to allow an efficient learning algorithm to output a  hypothesis representation $h$ which is  computationally very expensive to evaluate.  Thus, we shall only say that a PAC learning algorithm runs in time $t$ if any hypothesis it generates can be evaluated on any input $x \in X$ in time $t$ as well.  (This will be true for all of the positive results that we discuss in this chapter.)

The framework described above is often referred to as the ``distribution-free'' PAC learning model because of the requirement that a learning algorithm must succeed for any distribution ${\cal D}$. The original work \cite{Valiant:84} proposed several variants and extensions of this basic model, and a number of such variants have been extensively studied in the literature.  One natural variant is to restrict the model and only require learnability under certain particular types of probability distributions, such as the uniform distribution over $\zo^n$ or normal distributions over $\R^n$.  A different variant of the model (proposed already in \cite{Valiant:84}) is to allow the learning algorithm to make ``black-box''  queries on inputs $x \in X$ of its choosing. Yet another variant is the ``non-realizable'' setting, in which there need not be a perfect functional relationship between labels and the input instances with which they are paired (settings of this sort can also be viewed as allowing noise to be present in the labeling of input instances).  A range of interesting phenomena manifest themselves in each of these variants, and we will discuss each of them as well as some combinations of them later in this survey.

\subsubsection{An example of distribution-free PAC learning:  Learning Boolean conjunctions.} \label{sec:PAC-learning-example}

We instantiate the above framework by considering a concrete example of a PAC learning problem (learning Boolean conjunctions) and presenting an efficient PAC learning algorithm for this problem.  
This problem was already discussed in \cite{Valiant:84}, where an efficient algorithm was provided for it; our analysis here incorporates some subsequent refinements leading to slightly improved quantitative bounds, and introduces a tool (\Cref{claim:CHF} below) which will be useful for the subsequent discussion.

We take the domain to be $X=\zo^n$.  We consider the concept class ${\cal C}$, defined earlier, of all $3^n$ Boolean functions which are representable as Boolean conjunctions.
The following ``elimination algorithm'' works by eliminating those literals from the hypothesis conjunction which logically cannot be present in it given the labeled sample:

\begin{itemize}

\item Draw a sample $S$ of $m = O({\frac  1 \eps} \cdot (n + \log{\frac 1 \delta}))$ many labeled examples from $EX(f,{\cal D}).$

\item Initialize the hypothesis $h$ to be $x_1 \wedge \overline{x}_1 \wedge  \cdots \wedge x_n \wedge \overline{x}_n$, the conjunction of all $2n$ literals.
 For each \emph{positive example} $(x,1)$ in $S$, remove any literal $\ell$ from $h$ such that $\ell$ evaluates to 0 (false) under $x$.  (So, for example, if $n=4$ and there is a positive example $(x,1)$ in $S$ with $x=1100$, then the literals $\overline{x}_1,\overline{x}_2,x_3$ and $x_4$ would be removed from $h$.)

\item Output the resulting hypothesis $h$.

\end{itemize}

It is clear that the elimination algorithm has running time and sample complexity which are both $\poly(n,1/\eps,\log(1/\delta))$ (by the choice of $m$), and thus the algorithm is computationally and statistically efficient.  It remains to show that it is PAC, i.e. that for any unknown target conjunction $f$, with probability at least $1-\delta$ over the draw of the $m$-element sample $S$, the resulting hypothesis $h$ has error$_{\cal D}(h,f) \leq \eps$.

This turns out to be an easy consequence of a more general result about ``consistent hypothesis finders.''  A \emph{consistent hypothesis finder} for concept class ${\cal C}$ using hypothesis class ${\cal H}$ is an algorithm $A$ with the following property:  for all $m$, if $A$ is given a sample $S$ of $m$ labeled examples $(x,f(x))$ where $f$ is any fixed function in ${\cal C}$, then $A$ outputs a hypothesis $h \in {\cal H}$ which is consistent with $S$ (i.e. for each pair $(x,f(x))$ in $S$, the hypothesis $h$ satisfies $h(x)=f(x)$).  It is easy to see that the third step of the elimination algorithm corresponds to a consistent hypothesis finder for the class ${\cal C}$ of Boolean conjunctions using the hypothesis class ${\cal H}={\cal C}$.  Given this, the PAC property of the elimination algorithm (with the claimed value of $m$) follows directly from the following general result, which is proved via a simple union bound over ``bad'' hypotheses (hypotheses $h$ such that error$_{\cal D}(h,f) > \eps$):

\begin{claim} [\cite{BEH+:87}] \label{claim:CHF}
Let $A$ be a consistent hypothesis finder for a finite concept class ${\cal C}$ using a finite hypothesis class ${\cal H}$.  Fix any $f \in {\cal C}$ and any distribution ${\cal D}$, and let $h$ be a hypothesis obtained by running $A$ on a sample of
\[
m = {\frac 1 \eps} \left( \ln |{\cal H}| + \ln {\frac 1 \delta} \right)
\]
many independent examples from $EX(f,{\cal D})$.  Then 
with probability $1-\delta$ (over the random sample provided to $A$), the hypothesis $h$ satisfies error$_{\cal D}(h,f) \leq \eps.$
\end{claim}

Thus we have obtained the following theorem:

\begin{theorem} \label{thm:conjunction}
The class ${\cal C}$ of Boolean conjunctions is distribution-free PAC learnable in time $\poly(n,1/\eps,\log(1/\delta))$ using $\poly(n,1/\eps,\log(1/\delta))$ samples.
\end{theorem}

\medskip

\noindent {\bf Discussion:  On accuracy and confidence.} The logarithmic dependence on the confidence parameter $\delta$ which is achieved in the above result is a manifestation of a general phenomenon. As shown in  \cite{haukealitwar91}, given any PAC-type algorithm which can achieve an $\eps$-accurate hypothesis with confidence 0.9 (or 0.01), by independently running the algorithm $\ell = O(\log(1/\delta))$ times using accuracy parameter $\eps/2$ instead of $\eps$, with probability $1-\delta/2$ at least one of the hypotheses $h_1,\dots,h_\ell$ thus obtained will have accuracy at least $1-\eps/2$. By drawing a small additional number of labeled examples from $EX(f,{\cal D})$ it is possible to estimate the actual accuracy of each $h_i$ to within $\pm \eps/4$ with high confidence, and thus it is possible to obtain an $\eps$-accurate hypothesis with overall confidence $1-\delta$ with a running time and sample complexity dependence of only $O(\log(1/\delta))$ on the confidence parameter $\delta$. Given this, in our subsequent discussion of running times and sample complexities for various learning algorithms we will not mention the confidence parameter $\delta$. 

In fact, the theory of \emph{accuracy boosting} \cite{Schapire:90,Freund:95} shows that in the general distribution-free PAC learning model, for any learnable concept class it is always possible to achieve a dependence on the accuracy parameter $\eps$ which is polynomial in $1/\eps$.  This is much less obvious than the simple argument for the confidence parameter sketched above (in particular, it is not true in certain variants of the model such as the uniform-distribution PAC learning model discussed in \Cref{sec:distribution-specific}).   A rich theory of accuracy boosting has been developed over the years, encompassing many different algorithms (including the well-known AdaBoost algorithm of Freund and Schapire \cite{FreundSchapire:97}) and techniques.  Boosting now constitutes a well-developed and elegant theory, featuring deep connections to complexity theory, game theory, optimization, and other areas,  and is a highly influential export from computational learning theory to more applied areas of machine learning.  (For reasons of space this survey will not go more deeply into boosting, but see \cite{FreundSchapire:99short,Schapire2003,FreundSchapire12} for excellent overviews.)

Given the fact that accuracy boosting is possible in the distribution-free PAC model, in our descriptions of various distribution-free PAC learning results in \Cref{sec:distribution-free} we will also omit the dependence on the accuracy parameter $\eps$. Thus, for conciseness we could rephrase \Cref{thm:conjunction} as ``The class of Boolean conjunctions over $n$ variables is distribution-free PAC learnable in time $\poly(n).$'' (Note that giving an upper bound on the running time of a learning algorithm clearly implies the same upper bound on the sample complexity, since a time-$t$ learning algorithm can use at most $t$ examples.)

\medskip

\noindent {\bf Discussion:  On computational versus statistical aspects of efficient learnability.} \Cref{claim:CHF} highlights the importance of computational considerations in a theory of efficient learnability.  If ${\cal C}$ is any finite concept class, then taking ${\cal H}$ to be ${\cal C}$, a brute-force search algorithm (to find a consistent hypothesis in ${\cal H}$) will always yield a consistent hypothesis finder, and by \Cref{claim:CHF}, this means that any finite concept class ${\cal C}$ can be PAC learned with a sample complexity that is only logarithmic in $|{\cal C}|$. In other words, the number of samples required to learn any concept in ${\cal C}$ is proportional to the description length which suffices to specify any concept in ${\cal C}$.  In one sense this is a very powerful statement since it tells us, for example, that even as rich a class as ${\cal C} = \{$all $\poly(n)$-size Boolean circuits over $\zo^n\}$ can be PAC learned using only $\poly(n)$ many labeled examples.  But on the other hand, finding a consistent hypothesis from a prescribed class is often a computationally hard problem, as we will see in  \Cref{sec:representation-dependent}, and this is precisely why the simple \Cref{claim:CHF} is not a panacea for efficient learnability:  consistent hypotheses are powerful, but they may be hard to come by.

\subsection{The PAC learnability research program.} \label{sec:program}

Given the PAC learning framework that we have described above, a compelling meta-question suggests itself, which is to understand the running times and sample complexities that can be achieved for PAC learning various natural  concept classes of Boolean-valued functions.  The overarching goal is to determine which classes of Boolean functions are efficiently learnable and which are not.  Already in \cite{Valiant:84} Valiant identified this as a major potential line of  research:

\begin{quote}
``The results of \emph{learnability theory} would
then indicate the maximum granularity of the single
concepts that can be acquired without programming
\dots This paper attempts to explore the limits of what is learnable as allowed by algorithmic complexity \dots The identification of
these limits is a major goal of the line of work proposed
in this paper.''
\end{quote}
More recently, in his 2013 book \emph{Probably Approximately Correct: Nature's algorithms for learning and prospering in a complex world}, Valiant stated that 

\begin{quote}
``The questions of determining the most expressive classes of functions for which learning is still possible are the most fundamental questions of learning theory.''
\end{quote}

Indeed, in the years since \cite{Valiant:84}, a broad research program has been dedicated to this goal, and a wide range of positive and negative results on efficient learnability have been established in PAC-type frameworks.  
In this spirit, the main focus of the rest of this article is on surveying the progress that has been made towards mapping the boundary of efficient learnability in the distribution-free PAC model and some of its chief variants.  
Since the literature on this topic is vast, our coverage will  necessarily be incomplete; an attempt has been made to emphasize interesting and digestible techniques and concepts, sometimes at the expense of describing the technically most sophisticated known results.
We will see that while varying the learning model can change the scope of precisely which concept classes are efficiently learnable, there is a reassuring high-level robustness to the results that have been obtained:  more complex functions tend to be hard for efficient algorithms to learn, while simpler functions tend to be easy.

\section{Positive results:  PAC Learning Algorithms} \label{sec:algorithms}

In addition to defining the framework of PAC learning, \cite{Valiant:84} gave a number of results establishing efficient learnability of several interesting concept classes in the distribution-free PAC learning model and some of its variants.  We have already seen one such algorithm, for learning Boolean conjunctions, in \Cref{sec:PAC-learning-example}, and we will see some more algorithms from \cite{Valiant:84} below.  More broadly, since \cite{Valiant:84} a wide range of different learning results have been obtained, using many different techniques, across different variants of the PAC learning model.  In this section we will survey a number of these variants and algorithms, and discuss some of the key underlying ingredients.  

\Cref{sec:distribution-free} presents some of the main positive results in the distribution-free PAC learning model that was defined in \Cref{sec:key-ingredients}.  \Cref{sec:PAC-plus-MQ} surveys some of the major results in an augmentation of the distribution-free PAC learning model, the ``PAC + membership query'' model, in which the learning algorithm is allowed to make black-box queries to the target function in addition to receiving random labeled examples.  
\Cref{sec:noise} discusses several extensions of the basic distribution-free PAC model to allow for the possibility of having noisy data of various types.
Finally, \Cref{sec:distribution-specific} considers ``distribution-specific'' variants of the PAC learning model, in which the background distribution ${\cal D}$ is assumed to be some particular known ``benign'' distribution such as the uniform distribution over $\zo^n$ or the standard Normal distribution $N(0,1)^n$ over $\R^n$.  We will also consider combinations of the above models; for example, in \Cref{sec:distribution-specific} we will see both distribution-specific algorithms for learning with membership queries and distribution-specific agnostic learning algorithms.

\subsection{Distribution-free PAC learning from random examples.} \label{sec:distribution-free}

In this section we give an overview of some central algorithmic results for concept classes in the distribution-free PAC learning model that was defined in \Cref{sec:key-ingredients}. These results can be grouped into several general categories based on the core underlying algorithmic technique, and our presentation below follows this organization.

\medskip
\noindent {\bf The elimination algorithm for conjunctions and disjunctions.}  Recall the elimination algorithm from \Cref{sec:PAC-learning-example} for the concept class of all Boolean conjunctions over $\zo^n$.  It is easy to see that a dual algorithm is an efficient PAC learning algorithm for the concept class of all Boolean \emph{disjunctions} (which are ORs, rather than ANDs, of Boolean literals). This dual version uses negative examples to eliminate literals which logically cannot be present in the target disjunction.

As was already established in \cite{Valiant:84}, it is possible to learn significantly richer concept classes than just disjunctions or conjunctions with the elimination algorithm (though at the price of an increased running time) by using a simple yet powerful technique of \emph{feature expansion}, which plays an important role in many of the results discussed in this section.  To explain this, let us provide some definitions. A \emph{width-$k$ disjunction} is an OR of at most $k$ Boolean literals, and a \emph{$k$-CNF formula} is an AND of any number of width-$k$ disjunctions; for example, the function $f(x) = (x_4 \vee \overline{x}_6) \wedge (x_{1} \vee x_{8}) \wedge (\overline{x}_4 \vee \overline{x}_5) $ is a  2-CNF formula.  Since there are at most $N_k = (2n)^k$ many width-$k$ disjunctions over $n$ Boolean variables, by viewing each width-$k$ disjunction as a new ``meta-variable'' and running the elimination algorithm for conjunctions over an expanded feature space of $N_k$ meta-variables rather than the original $n$ Boolean input variables, we obtain a PAC learning algorithm for the class of $k$-CNF formulas running in time $\poly(N_k) = n^{O(k)}.$\footnote{Note that the distribution over the space of possible assignments to the $N_k$ meta-variables induced by this transformation is very different from the original distribution over the space of possible assignments to the $n$ underlying variables $x_1,\dots,x_n$ --- indeed, one is a distribution over $\zo^{N_k}$ while the other is a distribution over $\zo^n$. This is not a problem because of the PAC guarantee that the elimination algorithm for conjunctions succeeds under \emph{any} possible input distribution.}

Using the distributive law, it is easy to see that any $k$-term DNF formula (an OR of at most $k$ conjunctions of any width; for example, $f(x) = (x_1 \wedge x_2 \wedge x_3 \wedge \overline{x}_4) \vee (\overline{x}_3 \wedge x_5 \wedge \overline{x}_6)$ is a 2-term DNF) can be expressed as a $k$-CNF, and thus the above-described PAC learning algorithm for $k$-CNFs also PAC learns $k$-term DNFs, using $k$-CNF formulas as hypotheses.  Summarizing the above observations, we have

\begin{theorem}  [\cite{Valiant:84,PittValiant:86}] \label{thm:elimination-CNF}
For any $k$, the classes ${\calC}_{k\text{-}\CNF}$ and ${\cal C}_{\DNF(k)}$ of $k$-CNF formulas and $k$-term DNF formulas over $\zo^n$ can each be  distribution-free PAC learned in time $n^{O(k)}.$
\end{theorem}
By Boolean duality the same results hold for the classes ${\cal C}_{k\text{-}\DNF}$ and ${\cal C}_{\CNF(k)}$ of $k$-DNF formulas and $k$-term CNF formulas respectively.  

Information-theoretic arguments (which we mention briefly in \Cref{sec:hardness}) show that as long as $k$ is not too close to $n$, it is impossible for a PAC learning algorithm for ${\cal C}_{k\text{-}\CNF}$ to achieve $n^{o(k)}$ sample complexity (and hence of course also impossible to achieve $n^{o(k)}$ running time).  These information theoretic results do not apply to ${\cal C}_{\DNF(k)}$: it is conceivable that there is a PAC learning algorithm for $k$-term DNF formulas running in time $\poly(n,k)$.  However, four decades after the $n^{O(k)}$-time algorithm for $k$-term DNF sketched above was given in \cite{Valiant:84}, the $n^{O(k)}$ runtime of \Cref{thm:elimination-CNF} is still the best result known for constant $k$.  (We will see later that if the number of terms $k$ is sufficiently large, then faster running times than $n^{O(k)}$ are known.)

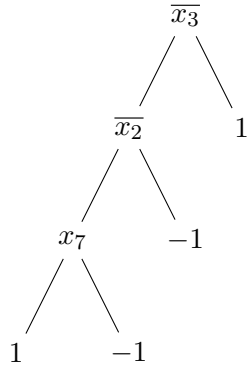
\begin{figure}[t]
\centering
\begin{tikzpicture}
level 1/.style={sibling distance=6cm},
level 2/.style={sibling distance=2.5cm}, 
level 3/.style={sibling distance=2.5cm}, 
  \node {$\overline{x_3}$}
    child { node {$\overline{x_2}$}
      child { node {$x_7$}
        child { node {$1$} }
        child { node {$-1$} } }
      child { node {$-1$} }
    }
    child {node {$1$}};
    \end{tikzpicture}
\caption{A 1-decision list ``if $\overline{x}_3$ then output 1 else if $\overline{x}_2$ then output $-1$ else if $x_7$ then output $-1$ else output $1$.''}
\label{fig:DL}
\end{figure}

\medskip
\noindent {\bf Greedy approaches for decision lists and decision trees.} A \emph{1-decision list} over $\{0,1\}^n$ is a nested sequence of ``if-then-else'' rules in which each ``if'' condition checks a single Boolean literal. For example, the decision list depicted in \Cref{fig:DL} is a 1-decision list of length 3 because three rules occur before the default output bit, which in this example is $1$.
A $k$-decision list is defined analogously but where now each ``if'' condition can be a conjunction of up to $k$ literals.  It is not difficult to show that the class of $k$-decision lists is a strict generalization of both the class of $k$-CNF formulas and the class of $k$-DNF formulas.

Rivest \cite{Rivest:87} introduced the class of decision lists and gave a simple greedy algorithm which is an efficient consistent hypothesis finder for the class of 1-decision lists using the hypothesis class of 1-decision lists.  
An efficient PAC learning algorithm for 1-decision lists follows from \Cref{claim:CHF}.  This extends to a learning algorithm for $k$-decision lists via the feature expansion technique described earlier in the context of $k$-DNF formulas (creating a new meta-variable for each length-$k$ conjunction), yielding the following:

\begin{theorem}  [\cite{Rivest:87}] \label{thm:k-DL}
For any $k$, the class ${\calC}_{k\text{-}\DL}$ of $k$-decision lists over $\zo^n$ can be distribution-free PAC learned in time $n^{O(k)}.$
\end{theorem}

\begin{figure}[t]
\centering
\begin{tikzpicture}[level/.style={sibling distance=60mm/#1}]
 \node {$x_3$}
    child { node {$x_5$}
      child { node {$x_1$}
        child { node {$1$} }
        child { node {$-1$} } }
      child { node {1} }
    }
    child { node {$x_1$}
      child { node {$x_2$}
        child { node {-1} }
        child { node {$x_4$}
        child { node {$1$} }
        child { node {$-1$} } } }
      child { node {$x_5$}
        child { node {$-1$} }
        child { node {$1$} } }
    };
    \end{tikzpicture}
    \caption{A decision tree of size eight and rank two.}
\label{fig:DT}
\end{figure}
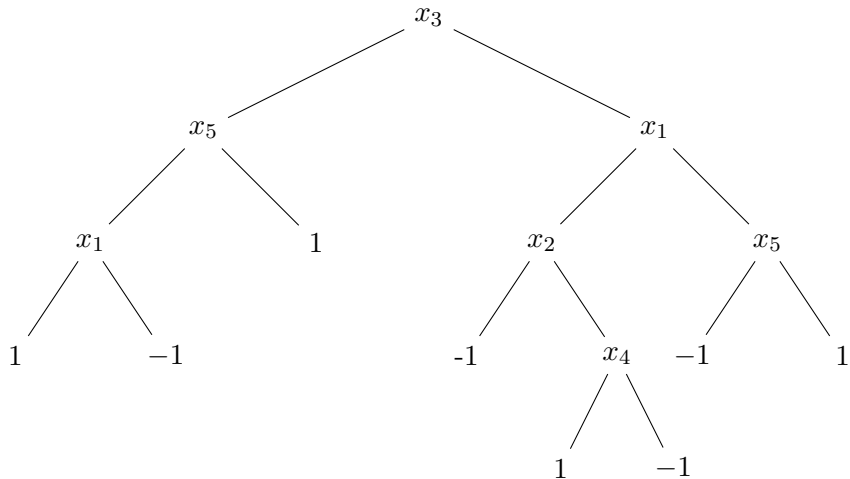

A closely related but more challenging concept class is the class of \emph{Boolean decision trees}.  A size-$s$ decision tree over $\zo^n$ is an $s$-leaf rooted full binary tree (each internal node has two childen) with an input variable $x_i$ labeling each internal node and an output bit labeling each leaf node.  The tree computes a Boolean function in the obvious way, by following the root-to-leaf path specified by the input assignment; for example, on input $10111$ the decision tree depicted in \Cref{fig:DT} would follow the rightmost path and have output 1.

Ehrenfeucht and Haussler \cite{EhrenfeuchtHaussler:89} introduced the notion of the \emph{rank} of a decision tree.  The rank of a leaf node is zero, and the rank of a tree $T$ with two subtrees $T_0$, $T_1$ is $\rank(T)=\max\{\rank(T_0),\rank(T_1)\}$ if $\rank(T_0) \neq \rank(T_1)$ and is $\rank(T)=\rank(T_0)+1$ if $\rank(T_0)=\rank(T_1)$.  Informally, $\rank(T)$ measures the maximum depth $r$ of a complete $2^r$-leaf binary tree which is contained in $T$; note that 1-decision lists correspond precisely to trees of rank 1.  Via a simple and elegant recursive algorithm which aligns with the recursive definition of rank given above, \cite{EhrenfeuchtHaussler:89} gave a consistent hypothesis finder for the class of rank-$r$ decision trees over $\zo^n$ using the same hypothesis class of rank-$r$ decision trees. Given a sample of $m$ examples that are labeled according to some rank-$r$ decision tree over $\zo^n$, their algorithm runs in time $O(m \cdot (n+1)^{2r}).$  Since any size-$s$ decision tree is easily seen to have rank at most $\log(s)$, this yields a quasi-polynomial time algorithm for learning decision trees:

\begin{theorem}  [\cite{EhrenfeuchtHaussler:89}] \label{thm:DT}
For any $s$, the class ${\calC}_{\DT(s)}$ of size-$s$ decision trees over $\zo^n$ can be distribution-free PAC learned in time $n^{O(\log s)}.$
\end{theorem}

Blum \cite{blu92} subsequently noted that Rivest's learning algorithm for $k$-decision lists recovers this result, taking $k=\log(s)$, by observing that any rank-$r$ decision tree can be expressed as an $r$-decision list. 
(We will see yet another algorithmic approach which recovers \Cref{thm:DT}, based on polynomial threshold functions, below.)

\medskip
\noindent {\bf Solving systems of linear equations.} Another algorithmic approach to PAC learning is based on solving systems of equations.  Here we work over $\F_2=\zo$ and we view $\zo^n$ as $\F_2^n$, observing that the AND of two elements of $\zo$ corresponds to multiplication over $\F_2$ and the XOR (parity) of two elements corresponds to addition over $\F_2$.

A \emph{parity} function over $\zo^n$ is a function $f(x)=\sum_{i \in S} x_i$ for some subset $S \subseteq [n]$, where addition is done over $\F_2$.  The class of all $2^n$ parity functions over $\zo^n$ corresponds to the class of all linear functions over $\F_2$-valued variables $x_1,\dots,x_n$. It is easy to see that a set of examples that have been labeled according to a  parity function corresponds to a consistent system of linear equations over $\F_2$, and thus Gaussian elimination can be used to solve the system and find a consistent parity function in polynomial time.  This yields an efficient PAC learning algorithm for the concept class of parity functions by \Cref{claim:CHF}. 

A generalization of the class of all parity functions is the class of all $\F_2$-polynomials of degree at most some fixed value $k$.
Using the familiar feature expansion technique of creating a new meta-variable for each monotone conjunction of up to $k$ Boolean variables, the above consistent hypothesis finder for parity functions extends to a consistent hypothesis finder for this class as well, and we obtain the following:

\begin{theorem}  [\cite{FischerSimon:92,HSW:92}] 
\label{thm:F2-polynomials}
For any $k$, the class of degree-$k$ $\F_2$ polynomials over $\zo^n$ can be distribution-free PAC learned in time $n^{O(k)}.$
\end{theorem}

\Cref{thm:F2-polynomials} can be viewed as giving a PAC learning algorithm for the class of parity-of-bounded-fanin-AND circuits.  A related result is that the concept class of AND-of-parity circuits is also PAC learnable in poly$(n)$ time  \cite{FischerSimon:92,HSW:92}.  This is achieved via a simple ``closure algorithm'' which is based on the observation that the set of points in $\zo^n$ that are satisfied by an AND-of-parity circuit corresponds to a vector subspace of $\F_2^n$, and hence the span of the positive examples is a vector subspace of $\F_2^n$ (corresponding to an AND-of-parity circuit) which is consistent with the sample.  
Since there are at most $2^{O(n^2)}$ many vector subspaces of $\F_2^n$ the claimed learning result follows from \Cref{claim:CHF}.

\medskip
\noindent {\bf Linear programming and learning linear threshold functions.}   A \emph{linear threshold function} (LTF) over $\R^n$ is a function $f(x)=\sign(\ell(x))$ where $\ell(x) = w_0 + w_1 x_1 + \cdots + w_n x_n$ is a real linear form and the function $\sign(\cdot)$ outputs $+1$ on non-negative inputs and $-1$ on negative inputs.  Linear threshold functions, also known as linear separators and halfspaces, play a fundamental and pervasive role throughout both theoretical and applied machine learning; indeed, they feature prominently in many of the learning results discussed throughout the rest of this article.

An example in $\R^n$ that is labeled according to a linear threshold function corresponds to a linear inequality in the unknown ``variables'' $w_0,w_1,\dots,w_n;$ for example, taking $n=3$, the labeled example $(-2.2, 1.4, 6; +1)$ corresponds to the inequality ``$ w_0 -2.2 w_1 + 1.4 w_2 + 6 w_3  \geq 0.$'' Thus the problem of finding a linear threshold function consistent with a sample translates to a linear programming feasibility problem, and known efficient algorithms for linear programming (e.g.~\cite{Kha:80,Karmarkar:84,Vaidya:96}) can be used to  find a feasible solution (a linear threshold function consistent with the examples) in polynomial time.\footnote{We remark that this discussion glosses over some subtle issues; one of these is that if the domain is $\R^n$, then a single coordinate of a single example is a real number which could in principle require infinitely many bits of precision.  Another is that the existence of a ``strongly polynomial time'' algorithm for linear programing is a famous open question; we refer the interested reader to \cite{Schrijver03} for a discussion of these and related issues.}

While the above discussion shows that there is a polynomial-time consistent hypothesis finder for the class ${\cal C}$ of $n$-variable linear threshold functions over $\R^n$ using hypothesis class ${\cal H}={\cal C}$, since this hypothesis class is infinite \Cref{claim:CHF} cannot be applied.  Fortunately, it turns out that a powerful generalization of \Cref{claim:CHF} provides the required tool for the job. Roughly speaking (see \cite{BEH+:89} for a precise statement) this result, which follows from the pioneering work of Vapnik and Chervonenkis \cite{VapnikChervonenkis:71} on uniform convergence of empirical probability estimates, essentially states that the ``$\ln |{\cal H}|$'' term in \Cref{claim:CHF} can be replaced with a combinatorial parameter of the hypothesis class ${\cal H}$ which is known as its \emph{Vapnik-Chervonenkis dimension}.  The Vapnik-Chervonenkis dimension of a class of Boolean functions ${\cal H}$ over domain $X$ (denoted $VC({\cal H})$) is the size of the largest subset of instances $S \subseteq X$ which is \emph{shattered} by ${\cal H}$, meaning that each of the $2^{|S|}$ many possible binary labelings of $S$ is achieved by some $h \in {\cal H}$.  Since, as is not too difficult to show,
the VC dimension of the class of linear threshold functions over $\R^n$ is $n+1$, we get the following fundamental result:  

\begin{theorem}  [\cite{BEH+:89}] 
\label{thm:LTF}
The class of linear threshold functions over $\R^n$ can be distribution-free PAC learned in time $\poly(n).$
\end{theorem}

\noindent {\bf Learning polynomial threshold functions and applications.}  Combining the feature expansion technique that we have seen earlier with the linear threshold function learning algorithm given by \Cref{thm:LTF} leads to a tool of unexpected power.  A \emph{polynomial threshold function} (PTF) of degree $d$ is a Boolean function $f(x)  =\sign(p(x))$ where $p(x_1,\dots,x_n)$ is a real polynomial of total degree at most $d$.  Since any degree-$d$ polynomial over $\R^n$ corresponds to a linear function over $O(n^d)$-dimensional space via our usual feature expansion technique, the VC dimension of the class of all degree-$d$ PTFs is at most $O(n^d)$. By running the linear programming based consistent hypothesis finder over this $O(n^d)$-dimensional space, we get the following corollary of \Cref{thm:LTF}:

\begin{theorem}  [\cite{BEH+:89}] 
\label{thm:PTF}
For any $d$, the class of degree-$d$ polynomial threshold functions over $\R^n$ can be distribution-free PAC learned in time $n^{O(d)}.$
\end{theorem}

An immediate consequence of \Cref{thm:PTF} is that it lets us automatically transform \emph{structural} results on representing different types of Boolean functions as low-degree polynomial threshold functions into \emph{algorithmic} results for learning the corresponding concept classes.  This turns out to be a powerful approach, yielding the fastest known running times for PAC learning algorithms for a number of concept classes over $\zo^n$.  

As a first illustration of the power of this technique, we observe that since any $k$-CNF or $k$-DNF formula is easily seen to be equivalent to some degree-$k$ PTF, \Cref{thm:PTF} immediately lets us recapture \Cref{thm:elimination-CNF}.  A second illustration is that since any $k$-decision list is also easily seen to be expressible as a degree-$k$ PTF, via Blum's observation (recall the discussion immediately after \Cref{thm:DT}) we recapture \Cref{thm:DT}.  While these two results have alternate proofs sketched earlier, for a number of other interesting and important concept classes the PTF approach is the only known method which achieves state of the art running times, as described below. 

\medskip

\noindent {\emph{DNF formulas and other Boolean formulas.}}  A longstanding goal, articulated already in \cite{Valiant:84} and reiterated in \cite{Valiant:85}, is to develop a $\poly(n)$-time algorithm for learning DNF formulas with $\poly(n)$ many terms. 
 (The $n^{O(k)}$ running time of \Cref{thm:elimination-CNF} for $k$-term DNFs is only $\poly(n)$ for constant $k$, and for any $k$ which is superlinear in $n$ it is larger than $2^n$ and hence trivial.)  Note that any $s$-leaf decision tree is easily seen to be expressible as an $s$-term DNF formula (simply by including a term for each 1-leaf in the decision tree)
 but the converse is not true,
 so learning DNF formulas is a more challenging problem than learning decision trees.  
 The first non-trivial algorithm for  DNF formulas with many terms was given by Bshouty \cite{bsh96}, who gave an algorithm that learns any $s$-term DNF over $n$ variables in time $2^{O(\sqrt{n \log s} \log^{3/2} n)}.$ At the
heart of Bshouty's algorithm is a structural result which shows that that any $s$-term DNF can be expressed as an $O((n \log n \log s)^{1/2})$-decision list (and hence as a PTF of degree $O((n \log n \log s)^{1/2})$). Tarui and Tsukiji \cite{TaruiTsukiji:99} subsequently gave a different algorithm for
learning DNF, which was based on the machinery of ``approximate inclusion/exclusion'' developed by
Linial and Nisan \cite{LinialNisan:90} in combination with hypothesis boosting;
their algorithm learns $s$-term DNF in time
$2^{O(n^{1/2} \log n \log s)}.$  
By combining ideas involving Chebyshev polynomials (implicit in the work of \cite{TaruiTsukiji:99}) and decision lists (from \cite{bsh96}), in \cite{KS04} it was shown that any DNF formula with $s$ terms can be expressed as a polynomial
threshold function of degree $O(n^{1/3} \log s)$. By \Cref{thm:PTF} this immediately yields the following learning algorithm for $\poly(n)$-term DNF (where the tilde-notation hides polylogarithmic factors):

\begin{theorem} [\cite{KS04}]
The class ${\calC}_{\DNF(s)}$ of $s$-term DNF formulas over $\zo^n$ can be PAC learned in time $2^{\tilde{O}(n^{1/3}  \log s)}.$
\end{theorem}

The PTF approach has also yielded non-trivial PAC learning results for Boolean formulas of \emph{arbitrary} depth and fixed polynomial size.  Perhaps surprisingly, quantum computing algorithms have played a crucial role in obtaining the necessary structural results.  Building on a line of work which gave query-efficient quantum algorithms for evaluating Boolean formulas \cite{ACRSZ10,FGG08,ReichardtSpalek08,Reichardt09}, Lee~\cite{Lee09formulas} showed that any $s$-leaf Boolean formula with AND, OR and NOT gates can be computed by a polynomial threshold function of degree $O(\sqrt{s})$. \ignore{(In fact, this work established a stronger property, namely that any size-$s$ Boolean formula can be \emph{approximately computed} --- so that the actual output value of the polynomial, and not just its $\pm 1$ sign, agrees with the output of the Boolean formula --- by a low-degree polynomial; we will revisit this stronger notion in \Cref{sec:agnostic}.)}  \Cref{thm:PTF} then yields the following:

\begin{theorem} [\cite{Lee09formulas}]
The class of all size-$s$ ($s$-leaf) Boolean formulas over $\zo^n$ can be PAC learned in time $n^{O(\sqrt{s})}.$
\end{theorem}

\begin{figure}[t]
\centering
\begin{tikzpicture}
\draw[gray, dashed] (-3,1) -- (2,-1);
\draw[gray, dashed] (-2,-1) -- (3,2);
\draw[gray, dashed] (-1,3) -- (2.5,-1);
\filldraw[black] (0,0) node{+};
\filldraw[black] (1,0.4) node{+};
\filldraw[black] (1.5,-0.4) node{+};
\filldraw[black] (0.7,0.2) node{+};
\filldraw[black] (1,2) node{$-$};
\filldraw[black] (1.5,1.3) node{$-$};
\filldraw[black] (-0.5,0.7) node{$-$};
\filldraw[black] (0,2) node{$-$};
\filldraw[black] (-0.7,1.5) node{$-$};
\filldraw[black] (-1.8,-0.5) node{$-$};
\filldraw[black] (-2.4,-0.2) node{$-$};
\filldraw[black] (2.4,1.2) node{$-$};
\filldraw[black] (1.6,0.6) node{$-$};
\filldraw[black] (0.4,-0.5) node{$-$};
\filldraw[black] (-0.4,-0.8) node{$-$};
\end{tikzpicture}
\caption{A collection of two-dimensional data points that have been labeled according to an intersection of three LTFs (the decision boundaries for the LTFs are depicted using dashed lines).}
\label{fig:intersection}
\end{figure}
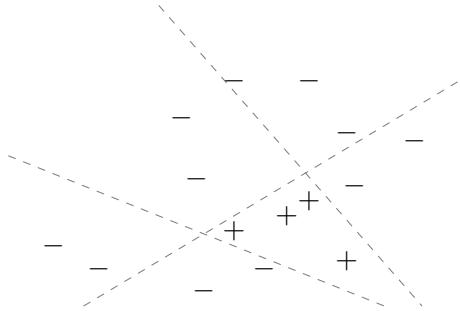

\noindent
\emph{Intersections 
of low-weight linear threshold functions.}  
Since linear threshold functions play such a prominent role throughout machine learning and computational learning theory, it is of significant interest to try to develop efficient PAC learning algorithms for functions that are defined in terms of several LTFs, such as the
intersection (AND) of two or more LTFs.  See \Cref{fig:intersection} for an illustration of a set of examples that have been labeled according to an intersection of three LTFs.  To date this effort has only been successful for functions defined in terms of LTFs that have ``low weight'' as described below.

The \emph{weight} of an LTF $f$ over $\zo^n$ is the smallest value $W$ such that $f$ can be expressed as $f(x) = \sign (w_0 + w_1 x_1 + \cdots + w_n x_n)$ where each $w_i$ is an integer and $\sum_{i=1}^n |w_i| = W.$  Well known results of Muroga \emph{et al.} \cite{MTT:61}, which were subsequently rediscovered by many authors,  show that every LTF over $\zo^n$ has weight at most $2^{O(n \log n)}$; complementing this upper bound, H\aa stad \cite{Hastad:94} has shown that there are $n$-variable LTFs which have  weight $2^{\Omega(n \log n)}.$  Known distribution-free algorithms for PAC learning intersections or other combinations of LTFs have a runtime dependence on the weight $W$, and are based on the PTF method as described below.

Using techniques of Beigel \emph{et al.} \cite{BRS:95} (which were in turn based on
classical low-degree rational function approximations, due to Newman \cite{Newman:64}, for the univariate functions $\sign(x)$), Klivans \emph{et al.} \cite{KOS:04} showed that any intersection of $k$ LTFs of weight $W$ is computed by a PTF of degree $O(k \log k \log W)$.
Via a different approach based on Chebyshev polynomials, it is also shown in \cite{KOS:04} that any intersection of $k$ LTFs of weight $W$ can be expressed as a PTF of degree $O(\sqrt{W} \log k)$; this gives a stronger bound in cases where $W$ is small and $k$ is large.  
These structural results directly yield the following PAC learning consequences via \Cref{thm:PTF}:

\begin{theorem} [\cite{KOS:04}]
 The class of intersections of $k$ weight-$W$ LTFs over $\zo^n$ can be PAC learned in time $n^{O(\min\{k \log k \log W,\sqrt{W} \log k\})}.$
\end{theorem}

\medskip
\noindent {\bf Limitations of the polynomial threshold function approach.} Since the polynomial threshold function approach to obtaining PAC learning algorithms is so effective, it is natural to try to understand its limits.  
There is an active line of work in computational complexity theory aimed at showing that various types of Boolean functions \emph{cannot} be expressed as low-degree polynomial threshold functions.  While structural results of this type do not absolutely rule out efficient PAC learnability of the concept classes in question, they suggest that new algorithmic techniques may be required to learn such classes.

The earliest lower bound results of this sort were given by Minsky and Papert \cite{MinskyPapert:68} in their celebrated work on ``perceptrons.'' (In our terminology, a ``perceptron'' is an LTF and a ``perceptron of order $d$'' is a degree-$d$ PTF.) They showed that the $n$-variable parity function  has polynomial threshold function degree exactly $n$, and that a particular $n^{1/3}$-term DNF formula (the ``one-in-a-box'' function) has polynomial threshold degree $\Theta(n^{1/3}).$  For both these results the upper bound is established by giving an explicit polynomial, and the lower bounds use a simple symmetrization argument to reduce the problem to one involving only univariate polynomials.

More recently, much more sophisticated techniques have been employed to show that other Boolean function classes do not have low-degree PTFs.  In \cite{Sherstov13sicomp,Sherstov13combinatorica} Sherstov showed that any PTF for the intersection of two weight-$n$ LTFs over $\zo^n$ may need to have PTF degree $\Omega(\log n)$, and that any PTF for the intersection of two general (high-weight) halfspaces may need to have PTF degree $\Omega(n)$.  Both lower bounds are optimal up to constant factors, the former by the results of \cite{KOS:04} mentioned earlier and the latter because any $n$-variable Boolean function trivially has PTF degree at most $n$.
These lower bounds use a range of  ingredients from polynomial and rational function approximation, Fourier analysis, and matrix perturbation theory.
 
 A related line of work 
 deals with PTF degree lower bounds for constant-depth circuits that are composed of AND, OR and NOT gates (where the AND and OR gates may have unbounded fan-in).  These are a natural generalization of DNF and CNF formulas, which can be viewed as depth-2 circuits of this type (the NOT gates are not considered as contributing to the circuit depth), and their learnability is of significant interest.  Despite much effort, to date no nontrivial distribution-free learning algorithms running in time $2^{o(n)}$ are known even for circuits of depth 3 (though see \cite{ST17itcs} for learning algorithms that achieve some  slight savings over a trivial $2^n$ runtime).   Building on a number of earlier results which use a broad spectrum of techniques \cite{Sherstov18breaking,Sherstov18power,BunThaler20}, the current state of the art is due to Sherstov and Wu \cite{SherstovWu19}, who showed that for any constant $k$, polynomial-size depth-$k$ circuits may require PTF degree $\tilde{\Omega}(n^{{\frac {k-1}{k+1}}})$. 

\subsection {\bf Distribution-free PAC learning with membership queries:  the ``PAC $+$ MQ'' model} \label{sec:PAC-plus-MQ}

This section deals with the basic distribution-free PAC learning model augmented with the ability to make black-box queries to the unknown target function $f$.  In such a query the learner specifies an arbitrary element $x \in X$ to a black-box oracle and receives as an answer the value $f(x) \in \zo$.  Such queries are often referred to in the literature as \emph{membership queries}, since the bit $f(x)$ indicates whether or nor $x$ belongs to the set $f^{-1}(1)$.

PAC learning with membership queries is a natural and well-motivated extension of the original PAC learning model.  There are many settings in which it may be possible to query the label $f(x)$ of a specific example $x$ of interest, such as when interacting with a teacher who can answer such queries, or when each example $x$ corresponds to a different scientific experiment that be designed and carried out by the learner.
As we shall see, the ability to make membership queries provides significant additional power and expands the classes of functions for which we know of efficient learning algorithms.

\medskip

\noindent {\bf Exact learning.}
The ``PAC $+$ MQ'' model is closely related to a well-studied \emph{exact learning} model which is known as the model of exact learning from membership and equivalence queries.  In this framework, which is suitable for finite domains such as $X=\zo^n$, there is no ambient probability distribution ${\cal D}$ over the domain $X$, and learning proceeds in a sequence of stages. At each stage the learner can either make a membership query as described above or an \emph{equivalence query}.  In an equivalence query the learner specifies a hypothesis $h: X \to \zo$, which corresponds to the question ``Is $h$ equivalent to the target function $f$ (in the sense that $h(x)=f(x)$ for every $x \in X$)?''  If $h$ is equivalent to $f$ then the learning process ends, and otherwise the learner is provided with a \emph{counterexample}, i.e. a point $x' \in X$ such that $h(x') \neq f(x').$  

An easy argument \cite{Angluin:88} shows that an equivalence query can be efficiently simulated in the PAC setting simply by drawing $O(\log(1/\delta)/\eps)$ many random labeled examples from the example oracle $EX(f,{\cal D}).$ If any of these labeled examples has $f(x) \neq h(x)$ then clearly $h \neq f$ and the required counterexample has been obtained; on the other hand, if $h$ is $\eps$-far from $f$ with respect to ${\cal D}$ (so the PAC learning success criterion has not been achieved) then the probability that $h$ and $f$ will agree on $O(\log(1/\delta)/\eps)$ many independent draws from ${\cal D}$ is at most $\delta$.  It follows that if a concept class ${\cal C}$ has an algorithm running in time $T$ in the model of exact learning from membership and equivalence queries, then it also has an algorithm running in time $\poly(T,1/\eps)$ in the PAC $+$ MQ model.  Indeed, nearly all of the learning algorithms in the PAC $+$ MQ model that we discuss below are derived from analogous algorithms in the exact learning framework.

As a proof of concept to demonstrate the power afforded by membership queries, we briefly consider the problem of learning an unknown \emph{$k$-junta}.  A Boolean function $f: \zo^n \to \zo$ is a $k$-junta if $f$ depends on at most $k$ of its $n$ input coordinates; for example, for $n=1000$, the function $f(x_1,\dots,x_{n}) = (x_1 \wedge \overline{x}_{375}) \vee (\overline{x}_1 \wedge x_{888})$ is a 3-junta.  Learning $k$-juntas in the PAC learning model without membership queries is a notoriously difficult open problem; despite intensive research effort and limited progress on the uniform-distribution restriction of the problem \cite{MOS04,Val15}, no algorithms with a faster than brute-force $n^k$ running time are known in the distribution-free PAC model discussed in \Cref{sec:distribution-free}. In contrast a straightforward approach, which crucially uses membership queries to carry out an efficient binary search in order to identify relevant variables, provides an algorithm in the model of exact learning from membership and equivalence queries that can learn any $k$-junta in time $\poly(n,2^k)$, and thus a PAC + MQ algorithm with the same running time.  

\medskip

\noindent
{\bf Monotone DNF formulas.}
A Boolean circuit or DNF formula is said to be monotone if it contains only AND and OR gates without any negations; thus a monotone DNF formula $F$ is simply an OR of terms $F(x) = T_1(x) \vee \dots \vee T_s(x)$ where each term is an AND of Boolean variables with no negations allowed. 
In the original PAC learning model where membership queries are not available, as shown by \cite{KLP+:87b} monotone DNF formulas are essentially no easier to learn than general DNF formulas. This is because given a DNF formula $F$ over literals $x_1,\dots,x_n,\ol{x}_1,\dots,\ol{x}_n$ and a distribution ${\cal D}$ over $\zo^n$, there is a corresponding monotone DNF formula $F'$ over $2n$ Boolean variables $x'_1,\dots,x'_{2n}$, obtained by replacing each un-negated literal $x_i$ with $x'_i$ and each negated literal $\ol{x}_i$ with $x'_{i+n}$, and a corresponding distribution ${\cal D}'$ over $\zo^{2n}$, where a draw of $\bx'=(\bx'_1,\dots,\bx'_{2n})$ from ${\cal D}'$ is obtained by drawing $(\bx_1,\dots,\bx_n) \sim {\cal D}$ and then setting $\bx'_i = \bx_i, \bx'_{i+n}=\ol{\bx}_i$ for each $i = 1,\dots,n.$  Learning the monotone DNF $F'$ under distribution ${\cal D}'$ corresponds precisely to learning the original DNF $F$ under distribution ${\cal D}$, so learning monotone DNF cannot be much easier than learning general DNF.

If membership queries are allowed, then the above reduction breaks down.  (As one way to see this, observe that in the context of the above reduction there are at most $2^n$ points in the support of distribution ${\cal D}'$ even though ${\cal D}'$ is a distribution over $\zo^{2n}$, but a membership oracle could query any of the $2^{2n}$ points in $\zo^{2n}$.)  This suggests that membership query algorithms may be more powerful than algorithms that can only use random examples for learning monotone DNF, and 
to the best of our current knowledge this seems to be the case. Indeed, already in \cite{Angluin:88}, by modifying an algorithm from \cite{Valiant:84} Angluin gave a simple and efficient algorithm for learning monotone DNF formulas in the model of exact learning from membership and equivalence queries.
The algorithm essentially builds the set of monotone terms in the target DNF formula one by one, where membership queries are crucially used to discover each successive term by ``walking down the hypercube.'' 
Translating to the PAC + MQ model, this yields the following:

\begin{theorem} [\cite{Valiant:84,Angluin:88}] \label{thm:mDNF}
The class of $s$-term monotone DNF over $\zo^n$ is learnable in the PAC $+$ MQ model in time $\poly(n,s).$
\end{theorem}

\medskip

\noindent
{\bf Exact learning other subclasses of DNF.}
Apart from monotone DNF, a wide range of polynomial time learning algorithms have been given for other subclasses of polynomial-size DNF formulas in the model of exact learning from membership and equivalence queries, and as explained earlier, all these results translate into PAC $+$ MQ learning algorithms.  These include polynomial-size Horn sentences (these are essentially DNF formulas in which at most one literal per term is negated; the learning algorithm for them, at a high level, can be viewed as an extension of the monotone DNF learning algorithm) \cite{AFP92}; read-twice DNF (these are DNF formulas in which each variable $x_i$ may occur at most two times) \cite{AizensteinPitt91,PR95} (see also \cite{Hancock:91}); and $\poly(n)$-leaf decision trees (via the ``monotone theory'' developed in \cite{Bshouty:95}, and via an approach based on learning generalized automata in \cite{BBB+:00}).  

A particular focus of research effort has been on developing efficient algorithms to learn unrestricted $k$-term DNF for $k$ as large as possible.  The current state of the art is that a range of different algorithmic approaches can exactly learn $k$-term DNF over $\zo^n$ in  time $\poly(n,2^k)$, but no faster running times are yet known. Blum and Rudich \cite{BlumRudich:95} were the first to achieve $\poly(n,2^k)$ runtime, through an algorithm based on randomized search; a different algorithmic approach is via the monotone theory of \cite{Bshouty:95}.  In contrast with these more sophisticated approaches, a relatively simple approach based on divide and conquer is given in \cite{Bshouty:97}, and another simple algorithm based on automata learning is given in \cite{Kushilevitz:97}.

\medskip

\noindent {\bf Exact learning beyond DNF formulas.} 
Efficient algorithms using membership and equivalence querise have also been given for several other natural concept classes.  A number of such results have been obtained for classes of read-once Boolean formulas (in which each input variable is allowed to occur at most once) composed of various kinds of allowed gates \cite{AHK93,BHH95,BHHK94} (see also \cite{HGM94}). Roughly speaking, the read-once condition ensures that a single variable has only a ``local'' effect on the target formula, and so membership queries may be used to probe  for structural information about the formula in the neighborhood of that unique occurrence.
A $\poly(n,s)$-time algorithm has also been given for the class of multivariate polynomials over $\F_2^n$ with at most $s$ monomials (equivalently, the class of parities of at most $s$ monotone conjunctions) \cite{SchapireSellie:96}.

Some of the most powerful techniques for exact learning with membership and equivalence queries are based on learning finite automata.  This approach was pioneered in the paper of Angluin \cite{Angluin:87}, which gave an algorithm that learns any unknown $m$-state deterministic finite automaton in time $\poly(m,n)$, where $n$ is the maximum length of any counterexample provided in response to an equivalence query.  This influential result led to a great deal of follow-up work; here we only mention the works \cite{BV96,OSK94,BBB+:00} on efficiently learning \emph{multiplicity} automata, which are a generalization of deterministic finite automata.  In \cite{BBB+:00} this approach was shown to yield efficient algorithms for many interesting concept classes, including decision trees, various subclasses of DNF formulas, parity of (not necessarily monotone) conjunctions over $\zo^n$ (extending \cite{SchapireSellie:96}) and other classes of multivariate polynomials, and more.

\medskip

We close this subsection by noting that while most PAC $+$ MQ learning algorithms are for concept classes over discrete domains such as $\zo^n$ and are based on exact learning, there are some exceptions.  For example, \cite{KwekPitt:98} give a PAC $+$ MQ learning algorithm for the class of intersections of $s$ LTFs over $\R^n$ that essentially\footnote{See \cite{KwekPitt:98} for a detailed explanation of the running time dependence on various parameters which arise because of the continuous domain $\R^n$, such as the minimum distance between the random examples and the separating hyperplanes defining the LTFs in the intersection.} runs in time $\poly(n,s,1/\eps)$ and is not based on exact learning (see also \cite{Baum:91b} for a related result).

\subsection{PAC learning in the presence of noise.}  \label{sec:noise}

As we have defined and discussed it so far, the PAC learning model assumes that all examples received by a learning algorithm are labeled correctly.
Of course, this may be an unrealistic assumption; thus it is natural to consider extensions of the basic PAC learning model which allow for the possibility of noise in the data.
Indeed, already in \cite{Valiant:85} Valiant proposed a variant of his original PAC learning model in which noisy examples may occur.

Many possible ways to incorporate noise into the PAC learning framework have been considered; in this subsection we briefly survey three of the most natural and widely studied noise models for PAC learning.
These are the \emph{malicious} noise model (in which adversarial noise can affect both instances and their labels), the \emph{agnostic} learning model (in which adversarial noise can affect only labels), and the relatively benign \emph{random classification noise} model (in which non-adversarial ``white noise'' affects only labels).

\medskip

\noindent 
{\bf Malicious noise.} One of the most difficult noise models for PAC learning is the so-called ``malicious'' noise model. In this model, first introduced in \cite{Valiant:85}, if the malicious noise rate is $\eta$ then independently each time the example oracle $EX(f,{\cal D})$ is invoked and an example $(\bx,f(\bx))$ is drawn with $\bx \sim {\cal D}$, with probability $1-\eta$ the learner receives the true example $(\bx,f(\bx))$ but with the remaining $\eta$ probability the learner receives an arbitrary pair $(x',b) \in X \times \zo$.  Since the pair $(x',b)$ is arbitrary, it should be viewed as being generated by an omniscient malicious adversary (hence the name of the model) whose goal is to thwart the learning algorithm.

Valiant \cite{Valiant:85} gave PAC learning algorithms for Boolean conjunctions and $k$-CNF formulas which can handle nonzero, but quite low, rates of malicious noise.  \cite{Valiant:85} mentioned the ``possibility that the learning phenomenon is only feasible with very low error rates,'' and this intuition was corroborated by the subsequent work of Kearns and Li \cite{KearnsLi:93}.
They showed that under a mild technical condition on the concept class ${\cal C}$, if the malicious noise rate $\eta$ exceeds $2\eps/(1+2\eps)$, then no PAC learning algorithm can achieve error rate less than $\eps$, regardless of the number of examples or amount of computation time it uses.
Thus $\eta = \Theta(\eps)$ is an information-theoretic upper bound on the malicious noise rate which $\epsilon$-error PAC learning algorithms can tolerate.

On the positive side, \cite{KearnsLi:93} established an extension of \Cref{claim:CHF} to the malicious noise model: any procedure for finding approximately consistent hypotheses from a fixed finite hypothesis class ${\cal H}$ yields a malicious noise tolerant PAC learning algorithm provided that the malicious noise rate is not too high (at most $\Theta(\eps)$).  \cite{KearnsLi:93} also gave a generic way to modify any polynomial-time PAC learning algorithm to handle some nontrivial level of malicious noise, essentially by first running the algorithm multiple times to generate a collection of candidate hypotheses and then using a fresh sample of random examples to select a high-accuracy final hypothesis from this pool of candidates. \cite{KearnsLi:93} show that this generic approach gives a polynomial-time PAC learning algorithm which can handle a malicious noise rate $\eta$ which is roughly $\log(s)/s$, where $s$ is the number of examples needed by the original polynomial-time PAC learner to achieve accuracy $\Theta(\eps)$ (with no malicious noise).
Since \cite{KearnsLi:93}, a number of works including \cite{KLS:09jmlr,ABL17,Shen21} have given PAC learning algorithms for linear threshold functions which can handle higher rates of malicious noise than $\log(s)/s$ for certain restricted classes of data distributions ${\cal D}$, such as uniform or log-concave distributions.  But in general, it seems quite challenging to give distribution-free malicious noise tolerant PAC learning algorithms for nontrivial concept classes which improve on the generic $\log(s)/s$ result of \cite{KearnsLi:93}.

\medskip

\medskip
\noindent
{\bf Agnostic learning.}
Another way of relaxing the strong assumptions of the standard (noise-free) PAC model is provided by the \emph{agnostic} learning model. This model was introduced by Kearns et al.~in \cite{KSS:94} and was based in part on the decision theoretic learning model studied by Haussler in \cite{Haussler:92}.  

In the agnostic PAC learning model, as in the standard PAC model, the labeled examples received by the learning algorithm are assumed to be  independently and identically distributed (note that this is not the case in the malicious noise model).  But unlike standard PAC learning, in agnostic learning each labeled example is a pair $(\bx,\by)$ drawn from an \emph{arbitrary} distribution ${\cal P}$ over labeled examples $X \times \zo$.  Thus the examples received by the learner are not assumed to be labeled by functions in any particular concept class, and indeed there need not be a functional dependence between the input instances $x$ and their associated labels $y$ at all, since the distribution ${\cal P}$ may have a nonzero probability of outputting both $(x,0)$ and $(x,1)$.  Agnostic learning is also sometimes referred to as ``non-realizable'' PAC learning; the original name for the model was chosen ``to emphasize the fact that as designers of learning algorithms, we may have no prior knowledge about the target function'' \cite{KSS:94}.

In agnostically learning a concept class ${\cal C}$ the crucial quantity (given the distribution ${\cal P}$) is the value $\mathrm{OPT}_{\cal C}$, which is the minimum possible prediction error of any concept $c \in {\cal C}$, i.e.,
\[
\mathrm{OPT}_{\cal C} = \inf_{c \in {\cal C}} \Prx_{(\bx,\by) \sim {\cal P}}[c(\bx) \neq \by].
\]
The goal of an ``agnostic learning algorithm for ${\cal C}$'' is to generate a hypothesis $h$ (which need not belong to ${\cal C}$) satisfying $\mathrm{error}(h,{\cal P}) \leq \mathrm{OPT}_{\cal C} + \eps$, where $\mathrm{error}(h,{\cal P}) = \Pr_{(\bx,\by) \sim {\cal P}}[h(\bx) \neq \by]$; in other words, the goal is to predict almost as well as the best concept in ${\cal C}$ (${\cal C}$ is sometimes referred to as a ``touchstone'' class in this context).
Agnostic learning can thus be viewed as learning ${\cal C}$ with a fixed distribution of adversarial noise in the labels that affects an $\mathrm{OPT}_{\cal C}$ fraction of the examples.

\cite{KSS:94} gave a range of results indicating that agnostic learning can be quite challenging; in particular, they showed that an efficient agnostic learning algorithm for conjunctions implies an efficient PAC learning algorithm for DNF. Only a few distribution-free polynomial-time agnostic learning algorithms are known, for fairly simple concept classes (though we will see later that more is possible for some natural fixed distributions). These include  certain one-dimensional ``piecewise'' concepts over the real line \cite{KSS:94}, certain types of decision trees (over non-Boolean domains) of depth at most two \cite{AHM95}, certain types of constant-dimensional geometric patterns \cite{GKS97}, and certain simple types of two-layer neural networks with bounded fan-in \cite{LBW96}.

In \cite{KKMS:08} Kalai et al.~gave a general method for agnostic learning which is similar in spirit to the ``PTF method'' described earlier for the original PAC learning model.
Recall that the PTF method shows that a fairly weak kind of polynomial approximation implies PAC learnability:  if, for every concept $f$ in a concept class ${\cal C}$ over $\zo^n$, there is a degree-$d$ polynomial whose sign agrees with $f$, then the class ${\cal C}$ is PAC learnable in time $n^{O(d)}$ via polynomial-time linear programming.
Kalai et al.~showed, using a polynomial-time algorithm for $L_1$ regression, that if every concept in ${\cal C}$ can be  approximated by a low-degree polynomial in a stronger sense, namely \emph{pointwise} approximation, then ${\cal C}$ is agnostically learnable:

\begin{theorem} [\cite{KKMS:08}] \label{thm:kkms}
Let ${\cal C}$ be a concept class over $\zo^n$ such that for every $f \in {\cal C}$, there is a degree-$d$ polynomial $p$ for which $|p(x)-f(x)|<\eps$ for all $x$.  Then ${\cal C}$ can be agnostically learned to error $\mathrm{OPT}_{\cal C}+\eps$ in time $\poly(n^d,1/\eps).$
\end{theorem}

As an application, \cite{KKMS:08} used \Cref{thm:kkms} to give a sub-exponential time algorithm for agnostically learning disjunctions (or conjunctions). Incorporating a sharp bound, due to Buhrman et al.~\cite{BCWZ:99}, on the degree of pointwise polynomial approximators for disjunctions in terms of the error parameter $\eps$, this result is
\begin{corollary}
[\cite{KKMS:08, BCWZ:99}]
The class ${\calC}$ of Boolean disjunctions (or conjunctions) is agnostically learnable to accuracy $\mathrm{OPT}_{\cal C}+\eps$ in time $n^{O(\sqrt{n \log(1/\eps)})}.$
\end{corollary}
 
\medskip
\noindent
{\bf Random classification noise.}
Finally, we consider the model of \emph{random classification noise} (RCN).  In this noise model, which was proposed by Angluin and Laird \cite{anglai88},  if the noise rate is $0<\eta<1/2$ then each example is drawn from $EX(f,{\cal D})$ but with probability $\eta$ the label is flipped and the negation of $f(\bx)$ rather than $f(\bx)$ is given to the learner as the label of $\bx$. 
Given the ``non-adversarial'' nature of the noise in this model, one might hope that relatively high noise rates can (at least sometimes) be tolerated, and this turns out to be the case.
Angluin and Laird proved a number of general results about this model, and in particular gave an efficient PAC learning algorithm for conjunctions in the presence of RCN at any rate $\eta < 1/2$ (the efficiency of the algorithm degrades inverse polynomially in the parameter $1/2 - \eta$, which can be shown to be unavoidable for any algorithm).  

While a handful of other RCN-tolerant learning algorithms were discovered for some specific concept classes in the years following \cite{anglai88}, a major advance came with the elegant work of Kearns \cite{Kearns:98}, which defined a new variant of the PAC learning model called the \emph{Statistical Query} (SQ) learning model. In this model the learning algorithm no longer receives individual random examples from $EX(f,{\cal D});$ rather, the learner interacts with a ``statistical query'' oracle which provides statistical information about the distribution of labeled examples. More precisely, a statistical query oracle is invoked by specifying a property of labeled examples (i.e.~a function $\phi: X \times \zo \to \zo$) and a tolerance parameter $0<\tau<1$, and in response the oracle returns a $\pm \tau$-additively accurate estimate of $\Ex_{\bx \sim {\cal D}}[\phi(\bx,f(\bx))]$, where $f$ is the target function being learned.  

It is straightforward to show that if a concept class ${\cal C}$ is efficiently learnable in the SQ model then it is also efficiently PAC learnable.  (Here ``efficient'' SQ learning meaning that ``not too many'' queries are made, that all query functions are efficiently evaluable, and that all tolerance parameters are ``not too small''). The significance of the SQ model for noise-tolerant PAC learning is that, as Kearns showed in \cite{Kearns:98}, any concept class ${\cal C}$ that is efficiently learnable in the SQ model is in fact also efficiently learnable in the PAC model \emph{even in the presence of random classification noise at any noise rate $\eta < 1/2$}. (As mentioned earlier for the \cite{anglai88} result, ``efficient'' RCN-tolerant PAC learning here allows for an (unavoidable) $\poly({\frac 1 {1/2 - \eta}})$ dependence on the noise rate $\eta$.)

This turns out to give a powerful and general method for designing RCN-tolerant PAC learning algorithms. 
\cite{Kearns:98}  showed that many known PAC learning algorithms, including most of the PAC learning algorithms described so far in this survey, can be shown to have SQ-model analogues, and thus the corresponding learning problems can be solved in the RCN model. Subsequent work showed that other PAC learning algorithms can be reworked to tolerate RCN using this approach as well; in particular, in \cite{BFK+:97}  Blum et al.~gave a new polynomial-time algorithm for learning linear threshold functions, and showed that this new algorithm can be recast in the SQ model.  This means that all of the PAC learning results obtained via the ``PTF method'', described earlier, extend to the RCN model.

Given the results described above it is natural to wonder whether \emph{every} concept class which is efficiently PAC learnable is also efficiently SQ learnable. Since SQ-learnable concept classes are PAC-learnable with RCN, this would establish an equivalence between PAC learning with and without RCN.  Already in \cite{Kearns:98} Kearns considered this question and gave a negative answer, by showing that there is no efficient SQ learning algorithm for the class of parity functions over $\zo^n$ (recall that, as described earlier, this class is efficiently PAC learnable in the standard model).  
The problem of learning parity with RCN is closely related to longstanding hard problems of decoding random linear codes and solving systems of noisy linear equations.  Despite much effort no efficient algorithm is known for  learning parity with RCN, and it is commonly believed that no polynomial-time algorithm exists.  (The current state of the art is a $2^{O(n/\log n)}$-time algorithm for noise rate $\eta$ any constant less than 1/2, due to Blum et al.~\cite{BKW:03}.)

We close this discussion of noisy PAC learning with a remark about the SQ model:  an attractive feature is that (as demonstrated by the negative result of \cite{Kearns:98} for the class of parities mentioned above) it is possible to prove \emph{unconditional} information-theoretic hardness results for various learning problems in the SQ framework.  A number of techniques have been developed for doing this, both for the usual ``high-accuracy'' learning notion that we have been considering throughout this survey and also for a notion of ``weak''  learning in which the goal is only to achieve accuracy slightly greater than $1/2$; see e.g.~\cite{bfjkmr94,Simon07,Szorenyi09,Feldman12}.
We further remark that in the decades since \cite{Kearns:98} the Statistical Query model has been shown to be of great relevance and utility in a number of related areas of theoretical computer science, including the study of \emph{differential privacy} \cite{BDMN05,DMNS06}, the study of adaptive data analysis \cite{DFHPRR15}, and the study of a computational model of \emph{evolution} that was introduced by Valiant \cite{Feldman08,Valiant09}.

\subsection{Distribution-specific PAC learning} \label{sec:distribution-specific}

The PAC learning model and its variants that we have discussed so far have all been ``distribution-free,'' since the learning algorithm must succeed for any possible (unknown and arbitrarily complex) distribution ${\cal D}$ over the domain $X$. 
An important strand of research in computational learning theory relaxes this requirement by considering \emph{distribution-specific} PAC learning; so now our learning algorithms will only be required to succeed in producing a high-accuracy hypothesis with high probability if the background distribution ${\cal D}$ is some specific, usually known (and usually highly structured), distribution, such as a product distribution over $\bn$ or a Gaussian distribution over $\R^n$.

Imposing a strong distributional assumption of this sort arguably runs contrary to the spirit of the original distribution-free PAC model, in which the distribution ${\cal D}$ was meant to capture an unknown and arbitrarily complex learning environment.  But on the other hand, distribution-specific PAC learning turns out to give rise to a host of interesting algorithmic problems with rich connections to other areas, and to allow for a much wider range of positive results and algorithmic techniques than the original distribution-free model.  For this reason, distribution-specific PAC learning has been the subject of intensive study.  

Of course, the first question in the study of distribution-specific learning is what distributions will be considered.  For learning over the domain $\bn$, product distributions (in which the values of all $n$ coordinates are mutually independent) are the most commonly studied type of fixed distribution.  
In this overview we will focus on learning under the uniform distribution over $\bn$, which is the simplest and most natural product distribution over the Boolean domain.  But we first make two remarks:  (1) In many cases, analogues of the uniform-distribution results we describe  can be established for broader classes of product distributions.  (2) There is also a rich body of work on distribution-specific learning over the continuous domain $\R^n$, with the two most intensively studied distributions being the uniform distribution over the unit sphere and the Gaussian distribution on $\R^n$.   This research has covered a wide range of topics including learning intersections of halfspaces
\cite{BlumKannan:97,Vempala:10}; agnostic learning of geometrically defined concepts including intersections of halfspaces, convex sets, sets with bounded Gaussian surface area,  and polynomial threshold functions \cite{KKMS:08,KOS:08,DKKTZ23}; and learning halfspaces in the presence of malicious noise \cite{KLS:09jmlr,ABL17}.  We also remark that some of these results have been extended to somewhat broader classes of distributions such as log-concave distributions over $\R^n$, see e.g.~\cite{KKMS:08,KKM:colt13,ABL17}. 

\medskip
\noindent
{\bf Uniform-distribution learning.}
One of the earliest papers to consider uniform-distribution learning over $\bn$ was due to Kearns, Li, Pitt and Valiant \cite{KLP+:87b}, which showed, among other results, that the class of \emph{read-once} DNF formulas (i.e.~DNF formulas in which each variable occurs at most once) is PAC learnable in $\poly(n,1/\eps)$ time under the uniform distribution (see also \cite{PagalloHaussler89}).  \cite{KLP+:87b} also showed, via a reduction, that in the distribution-free PAC model, read-once DNF are polynomial-time learnable if and only if the unrestricted class of all $\poly(n)$-size DNF is polynomial-time learnable.  Since distribution-free PAC learning of unrestricted DNF seems to be hard,  this was an early indication that uniform-distribution PAC learning can indeed be easier than distribution-free PAC learning.

In follow-up work, Kearns, Li and Valiant \cite{KLV:94} gave more evidence that uniform-distribution PAC learning is strictly easier than distribution-free PAC learning. \cite{KLV:94} considered the problem of learning arbitrary monotone Boolean functions (i.e.~functions for which flipping an input bit from 0 to 1 can never flip the output value from 1 to 0), and gave a $\poly(n)$-time algorithm that achieves error rate $1/2 - \Omega(1/n)$ under the uniform distribution (such a performance guarantee is often referred to as ``weak learning'').  In contrast, an easy information-theoretic argument shows that no $2^{o(n)}$-time algorithm can achieve such an error rate in the distribution-free model.

To gain some intuition for why uniform-distribution learning can be easier than distribution-free PAC learning, it is instructive to consider the problem of learning polynomial-size DNF formulas.  Under the uniform distribution, any term of length more than $t$ is satisfied by a uniform random input with probability only $2^{-t}$; it follows that for any $\poly(n)$-term DNF formula $F$, there is an $O(\log(n/\kappa))$-DNF formula $F'$ such that $F$ and $F'$ agree with probability $1-\kappa$ on uniform random inputs from $\zo^n$.  Building on this observation, \cite{Verbeurgt:90} proved the following theorem:

\begin{theorem} [\cite{Verbeurgt:90}] \label{thm:verbeurgt}
The class of all $\poly(n)$-term DNF formulas is PAC learnable under the uniform distribution in time $n^{O(\log(n/\eps))}$, using $\poly(n/\eps)$ many uniform random labeled examples.
\end{theorem}

In contrast with \Cref{thm:verbeurgt}, recall that for the distribution-free PAC learning model, no faster algorithm than $2^{\tilde{O}(n^{1/3})}$ is known for learning the class of $\poly(n)$-term DNF  (indeed, an arbitrary distribution can have high probability of satisfying a term even of length $\Theta(n)$, so the simple observation mentioned above does not hold for the distribution-free model).

A number of other early results used ad hoc techniques to give efficient uniform-distribution learning algorithms for various concept classes \cite{GKS93,Schapire94,HancockMansour:91,Hancock:93,GHM:96}.  A different approach to uniform-distribution learning was pioneered in the influential 1989 paper of Linial, Mansour and Nisan \cite{LMN89} (the journal version appeared in 1993 \cite{LMN:93}), based on \emph{Fourier analysis of Boolean functions}, which has since emerged as a very useful tool both for uniform-distribution learning and in many other branches of theoretical computer science (see \cite{ODonnell:book} and the many references therein).

\medskip
\noindent
{\bf Fourier analysis of Boolean functions and the low-degree algorithm.}
We  recall only the very basics; see the survey \cite{dewolf08} for a brief introduction or the book of O'Donnell \cite{ODonnell:book} for a detailed treatment. The starting point is the simple observation that any function $f: \bn \to \R$ has a unique expression as a multilinear polynomial.  The \emph{Fourier representation} of $f$ is simply this representation; i.e. writing $f: \bn \to \R$ as
\begin{equation} \label{eq:fourier}
\quad 
f(x_1,\dots,x_n) = \sum_{S \subseteq [n]} \hat{f}(S) \prod_{i \in S} x_i,
\end{equation}
the $2^n$ real coefficients $(\hat{f}(S))_{S \subseteq [n]}$ are the \emph{Fourier coefficients} of $f$.  (To motivate this terminology, observe that the $2^n$ parity functions $(\prod_{i \in S} x_i)_{S \subseteq [n]}$ form an orthonormal basis for the vector space of all real-valued functions over $\bn$, where the relevant inner product is $\langle f, g \rangle := \Ex_{\bx \sim \bn}[f(\bx)g(\bx)]$.) The \emph{Fourier $L_1$-norm} of $f$ is the sum of magnitudes $\sum_S |\hat{f}(S)|$ of $f$'s Fourier coefficients, denoted $\|\hat{f}\|_1$. 
Another crucial quantity is the \emph{Fourier $L_2$-norm} of $f$, written $\|\hat{f}\|_2$, which is equal to $(\sum_{S \subseteq [n]} \hat{f}(S)^2)^{1/2}$.  A simple argument using orthonormality of the basis functions gives that $\langle f, g \rangle =\sum_S \hat{f}(S) \hat{g}(S)$, and in particular, for any Boolean-valued function $f: \bn \to \bits$, \emph{Parseval's identity} says that $\|\hat{f}\|_2^2 = \langle f, f \rangle = 1.$ 

The paper of \cite{LMN:93} made two significant contributions. The first of these, the ``low-degree algorithm,'' uses the simple fact that each Fourier coefficient $\widehat{f}(S)$ is equal to $\Ex_{\bx \sim \bn}[f(\bx) \chi_S(\bx)]$, where $\chi_S(x)= \prod_{i \in S} x_i$ is the parity (basis) function over the variables in $S$. From this it can be shown that if every concept in ${\cal C}$ can be approximated by a low-degree polynomial in a fairly weak sense ($L_2$-approximation under the uniform distribution), then ${\cal C}$ is uniform-distribution PAC learnable from random examples. The learning algorithm works simply by estimating each of the low-degree Fourier coefficients and using them to build a real-valued hypothesis polynomial, which can be easily converted to a Boolean hypothesis:

\begin{theorem} [Low-degree algorithm \cite{LMN:93}] \label{thm:low-degree}
Let ${\cal C}$ be a concept class of $\bits$-valued functions over $\bn$ such that for every $f \in {\cal C}$, there is a degree-$d$ polynomial $p$ for which $\Ex_{\bx \sim \bn}[(p(\bx)-f(\bx))^2] \leq \eps$ (equivalently, $\sum_{|S| > d} \hat{f}(S)^2 <\eps$).  Then ${\cal C}$ can be PAC learned to error $O(\eps)$ in time $n^{O(d)}$.
\end{theorem}
It is interesting that, as was observed earlier in other settings, there is a close connection between a suitable notion of polynomial approximation and learnability.

The second contribution of \cite{LMN:93} was specific to the class $\mathsf{AC}^0$ of constant-depth, polynomial-size AND/OR/NOT circuits of unbounded fan-in. These circuits are of great interest in complexity theory and are a significant extension of polynomial-size DNF and CNF formulas (the depth-two case). Using an important result from complexity theory (H\aa stad's influential ``switching lemma'' \cite{Hastad:86}) and Fourier analysis of Boolean functions, \cite{LMN:93} showed that every $\mathsf{AC}^0$ circuit has almost all of its ``Fourier weight'' concentrated on degree at most $\polylog(n)$. Combining this with the low-degree algorithm, they gave a quasi-polynomial time algorithm for learning $\mathsf{AC}^0$:  

\begin{theorem} [\cite{LMN:93}] \label{thm:LMN}
The class of $\poly(n)$-size, depth-$d$ circuits over $\zo^n$ can be PAC learned in time $n^{O((\log (n/\eps))^d}.$
\end{theorem}
(See \cite{Hastad:01,Tal17} for some slight quantitative improvements on this bound.) 

A number of other uniform-distribution learning results have been obtained, based on the low-degree algorithm, by establishing Fourier concentration for various classes of functions.  A partial list of the concept classes for which such Fourier concentration results have been obtained includes the class of all ``juntas-of-LTFs'' (i.e.~Boolean functions $g(h_1,\dots,h_k)$ where $g$ is any $k$-variable Boolean function and the $h_i$'s are any LTFs) \cite{KOS:04};  various augmentations of $\mathsf{AC}^0$ which allow for a small number of LTF gates in addition to the usual AND and OR gates \cite{GS:10random,Kane14GL}; and degree-$d$ polynomial threshold functions \cite{DRST14,HKM14,Kane14GL}.

\medskip
\noindent {\bf The junta barrier.}
While the positive results described above capture some rather expressive classes of functions, on the other hand even fairly simple functions can resist the low-degree algorithm as far as $\poly(n)$-time learning is concerned.  As was discussed earlier, a function $f: \bn \to \bits$ is a \emph{$k$-junta} if it depends on only $k$ of the $n$ input variables, i.e.~$f(x_1,\dots,x_n)=g(x_{i_1},\dots,x_{i_k})$ for some $i_1 < \cdots < i_k$. Since $\poly(n)$-size DNFs, CNFs and decision trees can all compute $\log(n)$-juntas, obtaining $\poly(n)$-time learning algorithms for $\log(n)$-juntas is a prerequisite for polynomial-time learning of any of those richer classes. It is easy to see that a $k$-junta can compute the parity function over $k$ variables, and hence it may have no Fourier weight below level $k$, so the low-degree algorithm will not help learn such functions faster than time $\approx n^k$.
Indeed, it is a major challenge to give uniform-distribution learning algorithms for $k$-juntas that run much faster than time $n^k$; intuitively, the chief obstacle is how to identify the $k$ relevant variables faster than brute-force search.  Despite significant effort only a few positive results \cite{MOS04,Val15} are known, with the fastest current uniform-distribution junta learning algorithm, due to Gregory Valiant \cite{Val15}, running in time  $\approx n^{0.6k}$.

\medskip
\noindent {\bf Monotone functions.}
For monotone functions, it is easy to identify relevant variables by estimating degree-1 Fourier coefficients, and indeed it is folklore that monotone $k$-juntas can be learned in $\poly(2^k,n)$ time from uniform random examples.  Beyond juntas, learning different kinds of monotone functions under the uniform distribution has been the focus of much research.  Building on the aforementioned work of Kearns et al.~\cite{KLV:94}, several authors have studied the problem of learning arbitrary monotone Boolean functions. This line of work culminated with a Fourier-based algorithm, due to O'Donnell and Wimmer \cite{OW13}, which learns any monotone $f: \bn \to \bits$ to within error $1/2 - \Omega(\log(n)/\sqrt{n})$; this accuracy matches an information-theoretic lower bound, due to Blum et al.~\cite{BBL:98}, for $\poly(n)$-time algorithms learning monotone functions. On the other end of the spectrum of tradeoffs between runtime and error rate, Bshouty and Tamon \cite{BshoutyTamon:96} used the low-degree algorithm to show that any monotone Boolean function can be learned to error $\eps$ in time $n^{O(\sqrt{n}/\epsilon)}$, and a near-matching lower bound was given by Blais et al.~\cite{BCOST15}.  Several other classes of monotone functions have also been studied from the perspective of uniform-distribution learning, with the current state-of-the-art results for monotone DNF formulas \cite{Servedio:04iandc,Feldman12b} and monotone decision trees \cite{OdonnellServedio:07} being heavily based on Fourier techniques.

\medskip
\noindent {\bf Agnostic uniform-distribution learning.}
Researchers have also considered \emph{agnostic} uniform-distribution learning; in this framework the joint distribution ${\cal P}$ over $\bn \times \zo$ is promised to have its marginal distribution over $\bn$ be the uniform distribution.
The aforementioned work of Kalai et al.~\cite{KKMS:08}, shows that by using $L_1$ regression in place of the low-degree algorithm, every concept class ${\cal C}$ that can be learned using the low-degree algorithm (i.e.~every class of functions that have ``Fourier concentration,'' or equivalently, have low-degree-polynomial $L_2$-approximators under the uniform distribution) can in fact be \emph{agnostically} learned under the uniform distribution (with a slightly worse running time). Thus, the uniform-distribution learning results mentioned above which are based on Fourier concentration, including \cite{BshoutyTamon:96,KOS:04,GS:10random,Kane14GL,DRST14,HKM14}, all extend to uniform-distribution agnostic learning.

\medskip
\noindent
{\bf Uniform-distribution PAC learning with membership queries.}
{A significant body of research has also considered uniform-distribution PAC learning algorithms which are allowed to make membership queries. As we describe below, it turns out that membership queries are very useful for uniform-distribution learning;
indeed, we have already seen evidence of this in the gap between the $n^{\Theta(k)}$ fastest-known running time for learning $k$-juntas under the uniform distribution from random examples only \cite{Val15}, and the $\poly(n,2^k)$ running time that is achievable if membership queries are allowed (recall the discussion of junta learning in \Cref{sec:PAC-plus-MQ}).

An important tool for uniform-distribution membership-query learning was provided by Kushilevitz and Mansour \cite{kusman93}. Building on ideas of Goldreich and Levin \cite{GoldreichLevin:89}, who gave an efficient list-decoding algorithm for the Hadamard code in a cryptographic context, \cite{kusman93} gave a membership-query algorithm for efficiently identifying all  ``heavy'' (large-magnitude) Fourier coefficients of a Boolean function.
The key ingredient is a procedure (which makes crucial use of membership queries) to estimate the sum of squares of all Fourier coefficients which match a particular ``prefix'' of $\{1,\dots,n\}$, i.e.~the quantity 
\begin{equation} \label{eq:prefix}
\sum_{S \subseteq [n]:  S \cap [r] = T} \hat{f}(S)^2,
\end{equation}
where $r$ is any value in $\{1,\dots,n\}$ and $T$ is any subset of $\{1,\dots,r\}$.
Given this procedure, the main algorithm of \cite{kusman93} uses a ``branch-and-prune'' approach of 
considering increasing prefix lengths $1,2,\dots$ while maintaining a set of $r$-bit prefixes (corresponding to subsets $T \subseteq [r]$) for which (\ref{eq:prefix}) is estimated to be at least $\beta$. Parseval's identity ensures that 
the set of $r$-bit prefixes that are maintained at any stage cannot be of size larger than $O(1/\beta)$, which is crucial for the efficiency of the overall approach.  Using the fact that any coefficient $S \subseteq [n]$ such that $|\hat{f}(S)| \geq \beta > 0$ will always give rise to such a ``heavy'' prefix, it can be argued that this algorithm efficiently constructs a list which contains all ``heavy'' Fourier coefficients (of magnitude, say, at least $2 \beta$) and does not contain any ``light'' Fourier coefficients (of magnitude, say, less than $\beta/2$).

Given this ability to find ``heavy'' Fourier coefficients, a bit more analysis yields the following result for learning functions with small Fourier $L_1$ norm:

\begin{theorem} [\cite{kusman93}] \label{thm:KM}
The class of all functions from $\bn$ to $\bits$ which have Fourier $L_1$ norm at most $M$ is learnable, under the uniform distribution with membership queries, to accuracy $\eps$ in time $\poly(n,M,1/\eps).$
\end{theorem}

Kushilevitz and Mansour additionally proved that size-$s$ decision trees have Fourier $L_1$ norm at most $s$, and thereby obtained the following important corollary:

\begin{corollary} [\cite{kusman93}] \label{thm:KM}
The class of size-$s$ decision trees is PAC learnable, under the uniform distribution with membership queries, in time $\poly(n,s,1/\eps).$
\end{corollary}

We briefly mention a few even more powerful uniform-distribution membership-query learning results, proved using a range of different techniques.  Going beyond decision trees, Jackson \cite{Jackson:97} gave an efficient algorithm for learning DNF formulas: 

\begin{theorem} [\cite{Jackson:97}] \label{thm:jackson}
The class of $s$-term DNF formulas is PAC learnable, under the uniform distribution with membership queries, in time $\poly(n,s,1/\eps).$
\end{theorem}

The \cite{Jackson:97} algorithm combines the influential technique of \emph{accuracy boosting} \cite{Schapire:90,Freund:95} with the basic \cite{kusman93} algorithm, augmented to find ``heavy'' Fourier coefficients of certain real-valued functions which arise through the use of boosting.
(See also \cite{Feldman10} for a variant of the algorithm which uses an alternative approach based on agnostic boosting and does not involve finding Fourier coefficients of non-Boolean functions.)

Gopalan, Klivans and Meka \cite{GKM12colt} gave an algorithm which combines ideas from exact learning of branching programs using queries \cite{Angluin:87,BBB+:00} and complexity-theoretic work on approximating LTFs using branching programs \cite{MZ13} to efficiently learn combinations of LTFs with membership queries:

\begin{theorem} [\cite{GKM12colt}] \label{thm:GKM}
Let ${\cal C}_{LTF,any,k}$ be the class of all Boolean functions over $\zo^n$ which can be expressed as
$g(h_1,\dots,h_k)$ where each $h_i$ is a linear threshold function and $g$ is an arbitrary $k$-variable Boolean combining function.  The class ${\cal C}_{LTF,any,k}$ can be PAC learned with membership queries under the uniform distribution over $\zo^n$ in time $(n/\eps)^{O(k)}.$
\end{theorem}

In an intriguing result, Carmosino et al.~\cite{CIKK16} used a number of advanced ingredients from complexity theory, including the ``natural proofs'' of Razborov and Rudich \cite{RazborovRudich:97} and the Nisan-Wigderson pseudorandom generator \cite{NisanWigderson:94}, to give a quasi-polynomial time algorithm for learning $\mathsf{AC}^0[p]$, the class of constant-depth AND/OR/NOT circuits augmented with ``mod $p$'' gates:

\begin{theorem} [\cite{CIKK16}] \label{thm:CIKK}
For every prime $p \geq 2$, the class of size-$s$ $\mathsf{AC}^0[p]$ circuits is PAC learnable, under the uniform distribution with membership queries, in time $\exp(\poly(\log(ns/\eps)))$.
\end{theorem}

Finally, Gopalan, Kalai and Klivans \cite{GKK:08} combined Fourier techniques with tools from convex optimization to give an efficient \emph{agnostic} uniform-distribution membership-query learning algorithm for functions with bounded Fourier $L_1$ norm:
\begin{theorem} [\cite{GKK:08}]
Let ${\cal C}$ be a concept class over $\zo^n$ such that every $f \in {\cal C}$ has Fourier $L_1$ norm at most $s$.  Then ${\cal C}$ can be agnostically learned to error $\mathrm{OPT}_{\cal C}+\eps$, under the uniform distribution using membership queries, in time $\poly(n,s,1/\eps).$

\end{theorem}

Since, as mentioned earlier, size-$s$ decision trees have Fourier $L_1$-norm at most $s$, this gives an efficient uniform-distribution membership-query agnostic learning algorithim for decision trees.  (This result has subsequently been reproved using a number of different approaches, see \cite{KK09,Feldman10}.)

\section{Hardness of PAC learning} \label{sec:hardness}

In this section we discuss \emph{computational barriers} to efficient PAC learnability of different classes of Boolean functions.  The results which we describe are rigorous theoretical statements establishing formal limitations on what can be PAC learned by efficient algorithms, based on the presumed intractability of various computational problems.   (See the end of this introductory subsection for a brief discussion of why computational hardness assumptions of some sort are required.)

The negative results we present come in two general flavors. The first of these, which include the earliest negative results for PAC learning (originally given in work of Pitt and Valiant \cite{PittValiant:86,PittValiant:88}), are \emph{representation-dependent} statements which establish computational hardness for PAC learning algorithms that use a particular form of hypothesis representation, such as hardness of proper learning.  Results of this kind are generally based on NP-hardness; a suite of sophisticated techniques and powerful results have been employed in this area, and we briefly survey some of this work in \Cref{sec:representation-dependent}. The second strand, which is our main focus, consists of \emph{representation-independent} hardness results, which show that certain types of functions are \emph{inherently unpredictable} no matter what kind of efficiently evaluatable hypothesis representation a learning algorithm may use.  As we detail in \Cref{sec:representation-independent}, negative results of this sort were first obtained in the original work of Valiant \cite{Valiant:84}
and the subsequent work of Kearns and Valiant \cite{KearnsValiant:94}. Results of this sort were first based on hardness assumptions from cryptography; since, broadly speaking,
cryptography is about representing information in a way which makes it impossible for an adversarial observer to learn the information, this should perhaps not be a surprise.
However, more recent work has also fruitfully employed average-case hardness assumptions arising from complexity theory.

\medskip
\noindent {\bf On the need for assumptions to obtain computational hardness results.}  We briefly remark that our inability to prove strong lower bounds against general models of computation in complexity theory carries over to computational learning theory, and necessitates the use of unproven hardness assumptions, such as $\mathrm{P} \neq \mathrm{NP}$, in this area as well.  To see this, observe that if $\mathrm{P} = \mathrm{NP}$ then there would be a polynomial-time consistent hypothesis finder for any ``reasonable'' concept class (including the class of all unrestricted polynomial-size Boolean circuits), and hence by \Cref{claim:CHF} we would get a  proper polynomial-time learning algorithm for any such concept class.  Thus, in order to obtain  computational hardness results for learning ``reasonable'' concept classes, we must make computational hardness assumptions which are at least as strong as $\mathrm{P} \neq \mathrm{NP}$.

We remark in this context that many \emph{information-theoretic} lower bounds on learning are well known; indeed, the ``Fundamental Theorem of Statistical Learning'' (to use the terminology of \cite{SSBD}) states that the Vapnik-Chervonenkis dimension of a concept class characterizes, up to constant factors, the number of examples which are information-theoretically necessary and sufficient for PAC learning \cite{VapnikChervonenkis:71,BEH+:89}.  This result is easily seen to imply that even the broad class of all polynomial-size Boolean circuits can be PAC learned using a polynomial number of examples, albeit by an exponential-time algorithm. But as we will see in the rest of this section, it turns out that there are profound \emph{computational} obstacles to obtaining \emph{polynomial-time} PAC learning algorithms for rich concept classes.

\subsection{Representation-dependent hardness} \label{sec:representation-dependent} 
A \emph{representation-dependent} hardness-of-learning result is a statement of the following general type:  ``Assuming (some computational hardness assumption such as $\mathrm{NP} \neq \mathrm{RP}$), there is no polynomial-time PAC learning algorithm which uses hypothesis representations from class ${\cal H}$ to PAC learn concept class ${\cal C}.$'' 
The first results of this sort were established in the influential work of Pitt and Valiant \cite{PittValiant:86,PittValiant:88}. This paper gave a range of hardness results, including the following:  assuming that $\mathrm{NP} \neq \mathrm{RP}$, then (i) for each $k \geq 4$, the class of monotone $k$-term DNF is not PAC learnable in polynomial time using the hypothesis class of $(2k-3)$-term DNF; (ii) the class of monotone read-once (AND, OR, NOT) Boolean formulas\footnote{A formula is read-once if each input variable occurs at most once in it.} is not properly PAC learnable in polynomial time; (iii) the class of Boolean threshold functions\footnote{A Boolean threshold function is a linear threshold function $f: \zo^n \to \zo$ of the form $f(x)=\sign(u \cdot x - k)$, where each coordinate $u_i$ of $u$ belongs to $\{0,1\}$.} is not properly PAC learnable in polynomial time.  These proofs all work by giving reductions from known NP-complete problems to the problem of finding representations of concepts that are consistent with certain sets of labeled examples, where the representation must belong to the hypothesis class for which the hardness result is being shown.  

To make this more concrete, we give a simple example of such an argument by sketching a proof that if $\mathrm{NP} \neq \mathrm{RP}$, then there is no polynomial-time proper PAC learning algorithm for the class of 3-term DNF formulas.  (Our discussion below follows the argument  presented in Chapter~1 of \cite{KearnsVazirani:94}.) The proof is a reduction from the well-known NP-complete problem of \emph{graph 3-coloring}:  given as input an undirected graph, the problem is to determine whether it can be \emph{validly vertex-colored} (meaning that the two endpoints of every edge are assigned different colors) using a palette of only three colors.

The key to the reduction is a simple mapping $M$ which takes as input a graph with $n$ nodes and $m$ undirected edges (which is the input to the 3-coloring problem), and outputs a set $S = S^+ \cup S^-$ of $n$ positively labeled examples and $m$ negatively labeled examples. In more detail, for each vertex $i \in [n]$ there is a positive example which is the $n$-bit string with a 0 in only the $i$-th coordinate and 1's in all other coordinates, and for each edge $\{i,j\} \in {[n] \choose 2}$ there is a negative example  which is the $n$-bit string with 0s only in positions $i$ and $j$ and 1's in all other coordinates.  (See \Cref{fig:reduction} for an illustration of a possible input graph and the output of the mapping $M$ on this graph.)  The crucial property of this mapping, which is not difficult to show, is the following:  the graph $G$ has a valid vertex-coloring using three colors if and only if there is a three-term DNF formula $T_1 \vee T_2 \vee T_3$ which  labels all of the examples in $S^+$ as positive and all of the examples in $S^-$ as negative.  

Given this, the crux of the argument is to show that any polynomial-time proper PAC learning algorithm $A$ for the concept class of 3-term DNF formulas could be used to determine, in randomized polynomial time, whether or not an input graph can be validly vertex-colored using three colors.  The idea is to first perform the mapping $M$ on the input graph $G$ and then run the learning algorithm $A$, using the uniform distribution over $S^+ \cup S^-$  to simulate the example oracle $EX(c,{\cal D})$.\footnote{This is where random bits are used, and explains why the conclusion of the argument is about $\mathrm{NP}$ versus $\mathrm{RP}$ rather than $\mathrm{NP}$ versus $\mathrm{P}$.}
 Roughly speaking, if the graph has a valid 3-coloring then (with an appropriate setting of parameters) the PAC learning algorithm will with high probability construct a hypothesis 3-term DNF which correctly labels all of the examples in $S^+ \cup S^-$, whereas if the graph has no valid 3-coloring then since no such DNF exists the learning algorithm surely will not output one. Thus, by checking whether or not the output of the learning algorithm is a 3-term DNF consistent with $S^+ \cup S^-$, it is possible to efficiently and correctly (with high probability) determine whether or not the graph has a valid 3-coloring.

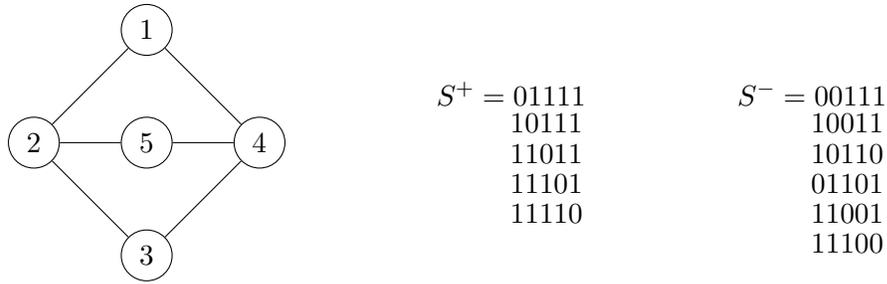
\begin{figure}[t]
\centering
\begin{tikzpicture}
\tikz[nodes={circle, draw}]  {
\node (1)  at (5,3) {1};
\node (2)  at (3.5,1.5) {2};
\node (5)  at (5, 1.5) {5};
\node (4)  at (6.5, 1.5) {4};
\node (3)  at (5, 0) {3};
\graph {
    (1)--(2)--(3)--(4)--(1); (2)--(5)--(4);
};
}
\filldraw[black] (3,2.5) node{$S^+=01111$};
\filldraw[black] (3.47,2.1) node{$10111$};
\filldraw[black] (3.47,1.7) node{$11011$};
\filldraw[black] (3.47,1.3) node{$11101$};
\filldraw[black] (3.47,0.9) node{$11110$};
\filldraw[black] (7,2.5) node{$S^-=00111$};
\filldraw[black] (7.47,2.1) node{$10011$};
\filldraw[black] (7.47,1.7) node{$10110$};
\filldraw[black] (7.47,1.3) node{$01101$};
\filldraw[black] (7.47,0.9) node{$11001$};
\filldraw[black] (7.47,0.5) node{$11100$};
\end{tikzpicture}
\caption{An input graph $G$ with $n=5$ nodes and $m=6$ edges, and the output $S^+ \cup S^-$ of the transformation $T$.}
\label{fig:reduction}
\end{figure}

\medskip
\noindent {\bf Representation-dependent hardness of learning and hardness of approximation.}  
In the simple argument sketched above, the hardness of properly learning 3-term DNFs is established using the NP-completeness of determining whether or not a graph has a valid 3-coloring.  Given this, it is should not be surprising that result (i) of \cite{PittValiant:88} mentioned earlier, namely the hardness of PAC learning $k$-term monotone DNF using $(2k-3)$-term DNF hypotheses, is established using the NP-hardness of \emph{approximating} the minimum number of colors needed to color a $k$-colorable graph.  This connection between computational hardness of representation-dependent learning and \emph{hardness of approximation}, which was first made in \cite{PittValiant:88}, turned out to be pivotal in subsequent work.  The 1990s and subsequent decades witnessed an explosion of research in hardness of approximation that resulted from the PCP theorem and the techniques employed in its proof;\footnote{Note that the early work of Pitt and Valiant \cite{PittValiant:86,PittValiant:88}  preceded these developments.} this progress in turn led to a wide range of different hardness results for representation-dependent learning, some of which are briefly summarized below.

One line of work gave representation-dependent hardness results in the original distribution-free PAC learning model.  Results of this sort show that under suitable complexity-theoretic hardness assumptions, various specific concept classes ${\cal C}$ are not PAC learnable in polynomial time using various specific hypothesis classes ${\cal H}$.
\cite{ABFKP:08} showed that DNF formulas and intersections of LTFs are not properly PAC learnable unless NP=RP, and that polynomial-size decision trees are not properly PAC learnable under a different hardness assumption.
\cite{KS08} showed that unless NP=RP, for any constant $k$ the class ${\cal C}_{\DNF(2)}$ of two-term DNFs cannot be PAC learned to accuracy $1/2 + 1/k$ in polynomial time using ${\cal C}_{\DNF(k)}$  hypotheses, and \cite{KhotSaket:11jcss} showed that under the same assumption, for any constant $k$ the class  of intersections of two LTFs cannot be PAC learned in polynomial time using the hypothesis class of arbitrary functions of $k$ LTFs.  Recently \cite{GS21} unified these two results by showing hardness of learning ${\cal C}_{\DNF(2)}$ using the hypothesis class of arbitrary functions of $k$ LTFs.
Even more recently, \cite{KST23} have shown that under the randomized exponential time hypothesis, no PAC learning algorithm for $s$-term DNFs, size-$s$ decision trees, or $\log(s)$-juntas can use DNF hypotheses and run in time $n^{\tilde{O}(\log \log s)}$ time (see  \cite{Bshouty23} for an extension to monotone versions of these classes).

A related strand of research proves hardness of representation-dependent \emph{agnostic} learning.
Strengthening an earlier result of \cite{BenEL03}, \cite{FGK+:06}
showed that if P $\neq$ NP, then even if there is an LTF that correctly classifies a $1-\eps$ fraction of a data set of labeled examples over $\R^n$, no polynomial-time algorithm can find an LTF that is correct on a $1/2 + \eps$ fraction of the examples, for any constant $\eps>0$ (i.e.~it is hard to agnostically properly learn LTFs even to 51\% accuracy). \cite{GR09journal} subsequently extended this hardness result to labeled examples over $\zo^n$, and \cite{FGRW09} extended it to the setting where there is a Boolean conjunction that correctly classifies $1-\eps$ fraction of the examples.
\cite{GKS10journal} established hardness of agnostically learning parities using low-degree $\F_2$-polynomial hypotheses, 
and \cite{DOSW:11,BGS18} established hardness of agnostically learning LTFs using low-degree PTF hypotheses.

Finally, a few papers have even established representation-dependent hardness of proper learning when membership queries are allowed.
\cite{Feldman:09jcss} extended the DNF hardness results of \cite{ABFKP:08} by showing that if NP $\neq$ RP then there is no polynomial-time proper PAC learning algorithm for DNF formulas even when the learning algorithm has access to an MQ oracle, and 
\cite{KST23b} extended the decision tree hardness results of \cite{ABFKP:08} by showing that under a plausible assumption about the exponential-time hardness of SAT, there is no polynomial-time proper PAC learning algorithm for decision trees even when the learning algorithm has access to an MQ oracle.

\subsection{Representation-independent hardness} \label{sec:representation-independent}
The representation-dependent hardness results discussed in \Cref{sec:representation-independent} provide useful insights for designers of learning algorithms, by giving strong evidence that  prospective algorithms which use various particular hypothesis classes are unlikely to be provably correct and efficient.
Even stronger evidence of hardness is provided by \emph{representation-independent} hardness results, which rule out the existence of provably correct and efficient learning algorithms that use \emph{any} kind of efficiently evaluatable hypothesis (under suitable hardness assumptions).  Three main approaches have been taken to establish such results, based on the cryptographic study of pseudorandomness; on public-key cryptography;  and, most recently, on average-case hardness assumptions arising from complexity theory.

\subsubsection{Pseudorandomness}
Already in \cite{Valiant:84} Valiant noted that the cryptographic notion of \emph{pseudo-random functions}, which had just recently been developed at that time in the seminal work of Goldreich, Goldwasser and Micali \cite{golgolmic86}, provides a  source of concept classes that are information-theoretically learnable from polynomially many examples but computationally hard to learn.  A \emph{pseudo-random function family} (PRFF) is a class ${\cal F}$ of functions $\{f_s: s \in \{0,1\}^n\}$, where each $f_s: \{0,1\}^n \to \{0,1\}$ is computable by a Boolean circuit of $\poly(n)$ size, with the following crucial property (roughly speaking): any $\poly(n)$-time distinguishing algorithm $A$ which is given black-box oracle access (i.e.~MQ access) to an unknown $f: \zo^n \to \zo$ cannot distinguish between the two possibilities that (i) $f$ is a truly random function drawn uniformly at random from all $2^{2^n}$ functions from $\zo^n$ to $\zo$, versus (ii) $f=f_s$ is a uniform random member of ${\cal F}$, i.e. a ``pseudo-random'' function.  Note that ${\cal F}$ is a set of ``only'' $2^n$ functions, each of which is efficiently computable. As Valiant observed, this definition implies that any PRFF ${\cal F}$ must be a concept class that is hard for any polynomial-time algorithm to learn, even under the uniform distribution and even given membership queries.  (This is because any polynomial-time learning algorithm for ${\cal F}$ would, with high probability, generate a high-accuracy hypothesis when run on $f \in {\cal F}$, but could not do so when run on a truly random function; since it is easy to estimate the accuracy of a hypothesis using a few random examples, such a purported efficient learning algorithm for ${\cal F}$ would easily yield a polynomial-time distinguisher between truly random and pseudo-random functions.) Since various widely accepted hardness assumptions, such as the existence of any one-way function \cite{HIL+:99}, are known to imply the existence of PRFFs, this shows that under such assumptions the class of all $\poly(n)$-size Boolean circuits is not efficiently PAC learnable even under the uniform distribution with membership queries.

The argument sketched above establishes that if ${\cal C}$ is a concept class which contains a pseudorandom function family, then ${\cal C}$ must be hard to learn.  
Naor and Reingold \cite{NaorReingold:04} proved that if factoring \emph{Blum integers} (integers of the form $N=pq$ where $p,q$ are primes congruent to 3 mod 4) is super-polynomially hard, then depth-5 circuits composed of Majority gates can compute pseudorandom functions and hence cannot be learned in polynomial time, even with membership queries and even under the uniform distribution.
Kharitonov had earlier \cite{kha93} obtained similar hardness-of-learning results, using similar assumptions, via an argument based on the existence of suitable pseudorandom generators; \cite{kha93} also showed, under a stronger but still plausible hardness assumption (that random Blum integers are hard to factor even in some sub-exponential time bound), that the quasi-polynomial running time of the \cite{LMN:93} uniform-distribution learning algorithm for $\mathsf{AC}^0$ circuits is close to best possible, even if membership queries are allowed.

A conceptually different approach, first explored in seminal work of Kearns and Valiant \cite{KearnsValiant:94}, turns out to establish hardness of learning for different, and in some cases even simpler, classes of Boolean circuits than those discussed above.  We turn to this approach, and to the work of \cite{KearnsValiant:94}, next.
 
\subsubsection{Public-key cryptography.}  

One of the central insights underlying \cite{KearnsValiant:94} is that the existence of secure \emph{public-key cryptosystems} (first proposed by Diffie and Hellman \cite{difhel76}) implies the hardness of learning problems associated with the decryption function of the cryptosystem.  
Recall that a public-key cryptosystem is a scheme that enables two parties, Alice and Bob, to communicate privately even in the presence of an eavesdropper Eve who can read the messages that are sent between them. The central notion enabling public-key cryptography is that of a ``trapdoor function;'' intuitively, this is an (encryption) function $f: \zo^n \to \zo^n$ that is polynomial-time computable by any party, but which is such that the (decryption) function $f^{-1}$ can only be computed in polynomial time by a party which is in possession of certain secret ``trapdoor'' information. 
(It is useful to think of the trapdoor function as being ``created'' by one of the parties, and of the trapdoor information as only being available to the creator of the trapdoor function.)
Trapdoor functions enable public-key cryptography\footnote{More precisely, secure communication from Bob to Alice; secure communication from Alice to Bob can of course be achieved simply by swapping the roles of the two parties.} in the following way:  Alice publishes an algorithm for computing a trapdoor function $f$ which she has created, thus making it possible for any party, including Bob, to compute $f$ given an input $x$.  To send a message $x$ to Alice, Bob simply computes $f(x)$ and sends the result to her.  Since Alice created the trapdoor function $f$ and is in possession of the secret trapdoor information, she can efficiently invert $f$ on the input $f(x)$ that Bob has provided and thereby recover the original message $x$; but the eavesdropper Eve, even though she is in possession of the encrypted message $f(x)$, lacks the trapdoor information which would enable her to invert $f$, so she is unable to obtain $x$.

The first crucial observation of \cite{KearnsValiant:94} is that the security of such a cryptosystem implies that \emph{the decryption function $f^{-1}$ must be hard to learn}. The reasoning is as follows: Since the eavesdropper Eve can compute $f$ on inputs of her choosing, she can create a data set of ``labeled examples'' of the form $(y,f(y))$.  Viewing each such pair $(y,f(y))$ as $(f^{-1}(x),x)$, this means that Eve can create a data set of examples labeled by the decryption function $f^{-1}$; if this decryption function were efficiently learnable, then Eve would be able to construct a high-accuracy hypothesis $h$ for $f^{-1}$, which violates the assumed security of the cryptosystem.  
Of course, a number of technical issues have to be dealt with in order to translate this high-level intuition into a formal proof (the trapdoor function $f$ must be a permutation or close to a permutation; we ultimately want hard-to-learn Boolean functions with single-bit outputs rather than $n$-bit-string outputs; etc), but the central idea is as sketched above.

\cite{KearnsValiant:94} showed that several different number-theoretic problems, each of which is conjectured to be hard in the average-case, can serve as the required hard-to-invert decryption functions in the framework proposed above. These include factoring Blum integers, inverting the RSA ``modular exponentiation'' function, and detecting quadratic residues.  Following \cite{KearnsVazirani:94}, we discuss (a slightly modified version of) the specific case of inverting the RSA function with exponent 3 in some detail, towards illustrating a second central idea of \cite{KearnsValiant:94}.  The encryption function $f$ is defined as $f(x)=f_N(x)=x^3 \mod N$, where $N=pq$ is the product of two $n$-bit primes  $p,q$ that are both congruent to  2 mod 3. The decryption function is $f_N^{-1}(y)=y^d \mod N$, where $d=(2(p-1)(q-1)+1)/3$; elementary number theory shows that $f_N$ is a bijection from $\Z^*_{N}=\{i \in \{1,\dots,N-1\}: \mathrm{gcd}(i,N)=1\}$ to itself, and that $f_N^{-1}$ is as defined above. Intuitively, the secret trapdoor information is the factorization $p,q$ of $N$; given this information it is easy to compute $f_N^{-1}$ in polynomial time. But a widely-accepted cryptographic assumption (the intractability of computing discrete cube roots) states that no $\poly(n)$-time algorithm can, given an $N$ which is the product of random $n$-bit primes  $p,q$ congruent to  2 mod 3 and a random input $y$, successfully compute $f_N^{-1}(y)$ with $1/\poly(n)$ success probability. Under this assumption, it can be shown that the concept class consisting of  all  functions  $f_N^{-1}(x)_i$ (the $i$-th bit of the $n$-output-bit function $f_N^{-1}$; note that every such functions is computable by a $\poly(n)$-size circuit) cannot be PAC learned in $\poly(n)$ time.
Hence, any class of Boolean circuits that is capable of performing modular exponentiation --- i.e. of computing functions of the form $x^d \mod N$ where $d,N$ are $O(n)$-bit numbers --- cannot be PAC learned in polynomial time. 

With the above sketch in hand, we now arrive at a second key insight  of \cite{KearnsValiant:94}, which is that by using ``encoding tricks'' to create an \emph{augmented} version of the input, it is possible to significantly reduce the circuit complexity of the decryption function while still maintaining the property that the decryption function is hard to learn.  In the context of the previous paragraph, computing the original decryption function $f_N^{-1}(y)=y^d \mod N$ using the standard approach of ``repeated squaring'' appears to require circuit depth $\Omega(n)$ (recall that $d=(2(p-1)(q-1)+1)/3$ is an $n$-bit-long integer).
The clever observation of \cite{KearnsValiant:94} is to consider, instead of $f_N^{-1}(y)=y^d \mod N$, a variant of this function which is given as its input not the string $y$ but an augmented input consisting of a list of successive squared powers of $y$, i.e.~
\begin{equation} \label{eq:augmented}
(y \mod N, y^2 \mod N, y^4 \mod N, y^8 \mod N, \dots, y^{2^{O(n)}} \mod N)
\end{equation}
(the output of the function is $y^d \mod N$ as before).  As Angluin put it in \cite{Angluin:92}, this amounts to ``mov[ing] some tasks that are computationally onerous but cryptographically irrelevant into the ``input''.''  (It is interesting to compare this trick with the ``feature expansion'' technique that was useful for obtaining many of the \emph{positive} results described earlier.)

The above-described variant function must still be hard to learn, because it is easy for any prospective learning algorithm which receives only the original input $y \mod N$ to efficiently construct the augmented input; so if there were an efficient learning algorithm for the variant function, then there would be an efficient learning algorithm for the original function.  (The input length of the variant function is quadratic in the original input length $|y|=O(n)$, but this is immaterial for polynomial-time algorithms.)  For a designer of lower bounds for learning algorithms, the considerable advantage of considering the variant function is that its circuit complexity is (to the best of our present knowledge) much lower than the circuit complexity of the original function.  This is because computing the variant function given \Cref{eq:augmented} as input only requires performing multiplication mod $N$ of a suitable subset of the input values $y ^{2^i} \mod N$, to build the desired exponent $d$ using its binary representation. Results of Beame, Cook and Hoover \cite{BCH86} show that this ``iterated multiplication mod $N$'' computation can be performed by a Boolean circuit of only $O(\log n)$ depth. Since such circuits are well-known to be equivalent to the complexity class $\nco$ of $\poly(n)$-size, $O(\log n)$-depth Boolean \emph{formulas}, this gives the following striking result:  under a suitable average-case hardness assumption (about factoring integers of the form $N=pq$ where $p,q$ are congruent to 2 mod 3), the concept class $\nco$ has no $\poly(n)$-time PAC learning algorithm.  Using similar ideas, and building on known reductions from circuit complexity and learning theory \cite{CSV84,PittWarmuth88,Reif87}, in \cite{KearnsValiant:94} it is also shown that under similar hardness assumptions, neither the the class of deterministic finite automata of $\poly(n)$ size, nor the class of constant-depth circuits composed of Majority gates, has a $\poly(n)$-time PAC learning algorithm.

A number of subsequent works have built on the ideas of \cite{KearnsValiant:94} to establish hardness of representation-independent learning, based on public-key cryptography, for even more limited concept classes or for even more powerful learning paradigms; we survey a few such results below. 

Angluin and Kharitonov \cite{angkha95} extended the basic \cite{KearnsValiant:94} hardness technique by showing that the existence of public-key cryptosystems \emph{secure against chosen ciphertext attack} implies hardness of learning for the corresponding decryption functions \emph{even if membership queries are allowed}.  Based on known public-key cryptosystems conjectured to be secure against chosen ciphertext attack, \cite{angkha95} extended several of the hardness results of \cite{KearnsValiant:94} to hold even against $\poly(n)$-time learning algorithms that can make membership queries.  \cite{angkha95} also showed, using the cryptographic notion of \emph{public-key signature schemes}, that under a plausible assumption about the existence of certain one-way functions, the class of $\poly(n)$-size DNF formulas is either $\poly(n)$-time PAC learnable without membership queries, or is not $\poly(n)$-time PAC learnable even if membership queries are allowed --- in short, ``membership queries won't help.''

Klivans and Sherstov \cite{KlivansSherstov09} extended the \cite{KearnsValiant:94} approach by analyzing lattice-based cryptosystems due to Regev \cite{Regev04,reg05} (see also \cite{FGK+:06} for related results).  
\cite{KlivansSherstov09} showed that the decryption function for Regev's cryptosystems can be computed as an intersection of $\poly(n)$ many degree-2 PTFs.  Using a simple feature expansion argument, it can be shown that an efficient learning algorithm for intersections of degree-2 PTFs yields an efficient learning algorithm for intersections of halfspaces.  Thus \cite{KlivansSherstov09} showed (among other results) that assuming that there does not exist a $\poly(n)$-time approximation algorithm (in a suitable approximation regime) for the ``unique shortest vector problem'' on lattices (which is the hardness assumption underlying the Regev cryptosystem),  for any constant $c>0$ the concept class of intersections of $n^c$ halfspaces over $\{0,1\}^n$ has no $\poly(n)$-time PAC learning algorithm.

More recently, Applebaum et al. \cite{ABW10} proposed new public-key cryptosystems based on the conjectured average-case hardness of various combinatorial problems, and provided some evidence for the average-case hardness of these problems. These public-key cryptosystems have decryption functions that can be computed by $O(\log n)$-juntas, and thus under the \cite{ABW10} average-case hardness assumptions, there is no $\poly(n)$-time PAC learning algorithm for the class of $O(\log n)$-juntas.

\subsubsection{Hardness based on average-case complexity assumptions.}  

The past decade or so has witnessed another exciting line of work on representation-independent hardness of learning, based on average-case hardness assumptions from complexity theory.  From a results perspective, this line of research is notable because it establishes computational hardness of learning various concept classes that consist of relatively simple functions, coming quite close to classes of functions which are known to be efficiently PAC learnable. From a techniques perspective, it is interesting that this new strand of research takes a different approach from prior work and is not based on the (presumed) intractability of computing the decryption function of a public-key cryptosystem. Instead, the key underlying hardness assumptions are that it is difficult to efficiently \emph{refute} certain particular types of unsatisfiable random constraint satisfaction problems.

To explain the general framework, we begin by briefly recalling some relevant background about the familiar constraint satisfaction problem (CSP) known as $k$-SAT.  
The $k$-SAT problem is based on the OR-predicate of arity $k$ (which we have been referring to as a ``clause'' of width $k$); a random $k$-SAT formula of size $m$ is obtained by ANDing together $m$ random clauses of width $k$ to form a random  $k$-CNF of size $m$.
In an influential paper, Feige \cite{Feige02} conjectured that if $m \geq Cn$ for a suitably large constant $C$, then random 3-SAT formulas of size $m$ are hard to \emph{refute}; this means, roughly speaking, that no efficient algorithm can distinguish such a random formula from a satisfiable formula of the same size.
(It is well known that for such large $m$, a random 3-SAT formula will with overwhelmingly high probability be unsatisfiable.) Feige \cite{Feige02} showed that this and related hardness assumptions, which are consistent with the current state of the art for known refutation algorithms, lead to surprising consequences in complexity theory and hardness of approximation.
In a similar spirit, the strand of work we are now discussing showed that assumptions of this flavor lead to strong representation-independent hardness-of-learning results.

The first result of this sort was due to Daniely, Linial and Shalev-Shwartz \cite{DLS14}. This work proposed a rather strong generalization of Feige's hardness-of-refutation assumption from \cite{Feige02}, called the ``strong random CSP assumption;” this assumption dealt with a range of non-standard predicates other than the 3-way OR-predicate, and assumed hardness of refutation even for values of $m$ much larger than $Cn$.  Under this new assumption, \cite{DLS14} established representation-independent hardness for a number of well-studied concept classes, including $\omega(1)$-size DNF formulas and intersections of $\omega(1)$ many halfspaces, as well as some results on representation-independent hardness of agnostic learning. The technical details of how \cite{DLS14} does this are beyond the scope of this article, but at a very high level, this is accomplished by relating the assumed hardness of refutation to the hardness of distinguishing between a sample of correctly labeled examples and a sample of randomly labeled examples.

As it turns out, soon after the appearance of \cite{DLS14} the specific ``strong random CSP assumption'' of \cite{DLS14} was shown to be false by Allen et al.~\cite{AOW15}, who gave an efficient refutation algorithm for this particular problem.
However, the general approach introduced by \cite{DLS14} turned out to be quite adaptable to using different assumptions, and (at the time that this article was written) this general approach is currently the source of quite a number of state-of-the-art representation-independent hardness of learning results, based on assumptions that have withstood considerable scrutiny, as briefly surveyed below.

Daniely et al \cite{DSS16} showed that no $\poly(n)$-time algorithm can learn $\omega(\log n)$-term DNF formulas, using an assumption that is considerably weaker and more standard than the ``strong random CSP assumption'' of \cite{DLS14}. The assumption of \cite{DSS16}, like that of Feige \cite{Feige02}, only involves the OR-predicate; it is that given any $d$, there exists some large enough value $k(d)$ such that random $k(d)$-SAT instances are hard to refute. \cite{DSS16} further showed that this same assumption implies the hardness of various other learning problems, including learning intersections of $\omega(\log n)$ many halfspaces, agnostically learning conjunctions, halfspaces, or parities, and learning sparse PTFs.  The DNF and intersection-of-halfspaces hardness results of \cite{DSS16} were subsequently strengthened by Daniely and Vardi \cite{DV21} to hold for $\omega(1)$-term DNF and intersections of $\omega(1)$ many halfspaces, under a slightly different hardness assumption.

In \cite{Daniely16halfspace}, Daniely similarly eliminated the need for the non-standard ``strong random CSP assumption'' of \cite{DLS14} to achieve hardness results for agnostically learning a single halfspace. He used a hardness assumption based on the $k$-XOR predicate (corresponding to parity of $k$ variables), and showed that if random $k$-XOR formulas of sufficiently large size are hard to refute, then for any arbitrarily small constant $\eta>0$, no polynomial-time algorithm can find a hypothesis whose accuracy is even $1/2 + 1/\poly(n)$, given examples from a joint distribution ${\cal D}$ over $\R^n \times \{-1,1\}$ for which some halfspace correctly labels a $1-\eta$ fraction of examples from ${\cal D}$.

\section{Conclusion}
We have seen how Valiant's seminal paper \cite{Valiant:84} gave rise to a rich and multifaceted theory of computationally efficient learning, which is being actively investigated to this day. Extending beyond the scope of this article, in the decades since \cite{Valiant:84} the PAC learning framework has grown to encompass a broad range of topics and questions that go beyond the learnability and non-learnability of specific classes of Boolean functions. 
Within the realm of computational learning theory, these include, among many other topics, the study of \emph{accuracy boosting} methods (which, as mentioned in \Cref{sec:PAC-learning-example}, are widespread and useful in computational learning); PAC-style models of learning for non-Boolean valued functions; and PAC-style learning in \emph{unsupervised} scenarios where examples are unlabeled.  
Variants of the PAC learning model have played a central role in investigating a variety of other phenomena in learning, including privacy, active learning, memory-sample tradeoffs, streaming, and reproducibility.
Rich connections have also been established between the PAC learning framework and many other subfields of theoretical computer science apart from learning, including but not limited to property testing, algorithmic game theory, computational complexity, statistical algorithms, and cryptography.  Zooming out even more, ``PAC-style thinking'' has had a tremendous influence on the development of machine learning over the past forty years, and has served as a source of conceptual scaffolding for the study of phenomena such as neural computation and evolution \cite{Valiant00,Valiant09,ValiantPAC}.  It will be exciting to see where the next forty years of reflection on the ideas from \cite{Valiant:84} take us.

\section*{Acknowledgements}
I thank the referees for their very helpful comments and corrections, which greatly improved this chapter.

\begin{footnotesize}
\bibliography{a}{}
\bibliographystyle{alpha}
\end{footnotesize}
\end{document}